\documentclass[10pt,compsoc,journal]{IEEEtran}
\usepackage{algorithm}
\usepackage{algorithmic}
\usepackage{amsfonts}
\usepackage{amsmath}
\usepackage{amssymb}
\usepackage{amsthm}
\usepackage{array}
\usepackage{bbding}
\usepackage[backend=biber,defernumbers,isbn=false,sorting=none,style=numeric,url=false]{biblatex}
\usepackage{bm}
\usepackage{booktabs}
\usepackage[labelsep=period]{caption}
\usepackage{color}
\usepackage{colortbl} 
\usepackage[ddmmyyyy]{datetime}
\usepackage{enumitem}
\usepackage{flushend}
\usepackage[bottom]{footmisc} 
\usepackage{graphicx}
\usepackage{hyperref}
\usepackage[misc]{ifsym}
\usepackage[switch]{lineno} 
\usepackage{mathrsfs}
\usepackage{mathtools}
\usepackage{multirow}
\usepackage{nicematrix}
\usepackage{ragged2e}
\usepackage{siunitx}
\usepackage{stfloats}
\usepackage[caption=false,font=normalsize,labelfont=sf,textfont=sf]{subfig}
\usepackage{tabularx}
\usepackage{textcomp}
\usepackage{tikz}
\usepackage{tipa} 
\usepackage[table]{xcolor}
\usepackage{verbatim}

\renewcommand{\figurename}{Figure}
\renewcommand{\tablename}{Table}

\newcolumntype{P}[1]{>{\centering\arraybackslash}p{#1}}
\newcolumntype{Y}{>{\centering\arraybackslash}X}

\graphicspath{{Figure_folder/}}

\DeclareSourcemap{
  \maps[datatype=bibtex,overwrite]{
    \map{
      \perdatasource{references-SI.bib}
      \step[append,fieldset=keywords,fieldvalue=web]
    }
  }
}

\begin{document}


\title{Topology-enhanced machine learning for speech signal processing}

\author{
    Pingyao Feng\textsuperscript{1,2,$\ast$}, Qingrui Qu\textsuperscript{1,3,$\ast$}, Haiyu Zhang\textsuperscript{1}, Siheng Yi\textsuperscript{1}, Zhiwang Yu\textsuperscript{1,4}, Zeyang Ding\textsuperscript{1}, Yifei Zhu\textsuperscript{1,$\dagger$} 



    \thanks{
        \em \textsuperscript{1}Department of Mathematics, Southern University of Science and Technology, Shenzhen, China.  
        \textsuperscript{2}Present address: Department of Mathematics, North Carolina State University, Raleigh, NC, USA.  
        \textsuperscript{3}Present address: School of Intelligence Science and Engineering, Harbin Institute of Technology (Shenzhen), Shenzhen, China.  
        \textsuperscript{4}Present address: School of Mathematics and Systems Science, Shenyang Normal University, Shenyang, China.  \\
        $\ast$ These authors contributed equally: Pingyao Feng, Qingrui Qu \\
        $\dagger$ Corresponding author: \href{mailto:zhuyf@sustech.edu.cn}{zhuyf@sustech.edu.cn} 
    }
}

\maketitle

\begin{abstract}
    In artificial-intelligence-aided signal processing, existing deep learning models often exhibit a black-box structure.  Here, conceptually beyond spectral analysis, we demonstrate that topological methods not only effectively capture intrinsic and complex structural information but can also enhance neural networks.  We provide a transparent methodology, TopCap, to capture topological features inherent in time series for basic machine learning.  Compared to prior approaches, we obtain descriptors that probe finer information such as the vibration of a time series.  Notably, in classifying voiced and voiceless consonants, TopCap achieves an accuracy consistently standing in comparison with neural network models.  Moreover, by integrating TopCap features into those neural networks, our approach improves upon state-of-the-art methods in terms of robustness against noise, as well as accuracy, stability, convergence of loss function, and interpretability.  
\end{abstract}


\section*{Introduction}

\IEEEPARstart{I}{n} 1966, Mark Kac asked the famous question: `Can you hear the shape of a drum?'  To hear the shape of a drum is to deduce information about the shape of the drumhead from the sound it makes, using mathematical theory.  In this article, we venture to flip and mirror the question across senses and address instead: `Can you see the sound of a human speech?'  

As a major task of natural language processing (NLP), speech recognition is one of the essential components of artificial intelligence (AI).  In turn, AI advancements have led to a widespread adoption of voice recognition technology, encompassing applications such as speech-to-text conversion and music generation.  The continuing growth of topological data analysis (TDA) \cite{carlsson_topology_2009} has integrated topological methods into many areas including AI \cite{Gabrielsson2019,ML2,DL3,tdl}, which makes neural networks (NN) more interpretable and efficient, with a focus on structural information beyond the point-set level \cite{baez}.  

In the field of voice recognition \cite{speech_recog1,speech_recog2}, more specifically consonant recognition \cite{consonant3,consonant4,consonant1,consonant5,consonant2}, prevalent methodologies frequently revolve around the analysis of energy and spectral information, which may be viewed as biomimetic engineering (see Supplementary Sec.~\ref{sec:phone}).  While topological approaches are still rare in this area, we combine TDA to machine learning (ML) and obtain a classification for speech data, based on geometric patterns hidden within phonetic segments.  The method we propose, TopCap (referring to the capability of capturing topological structures of data), is applicable not only to speech data but also to related time series data that require extraction of structural information for ML algorithms.  Moreover, it implements state-of-the-art NNs to generate their topology-enhanced counterparts TopNNs.  
\\

\noindent
Conceptually, TDA is an approach that examines data structure through the lens of topology.  This discipline was originally formulated to investigate the shape of data, particularly point-cloud data in high-dimensional spaces \cite{persistent_shape}.  Characterised by insensitivity to metrics, robustness against noise, invariance under continuous deformation, and coordinate-free computation \cite{carlsson_topology_2009}, TDA has been combined with ML algorithms to uncover intricate and concealed information within datasets \cite{Gabrielsson2019,ML2,DL3,ML4,ML6,ML1,ML5}.  In this context, topological methods have been employed to extract structural information from the datasets, which in turn enhances the efficiency of the original algorithms.  Notably, TDA excels in identifying patterns such as clusters, loops, and voids in data, which distinguishes it as a burgeoning tool in the realms of data science and AI \cite{introDS}.  With its distinctive emphasis on the shape of data, TDA has led to applications in various far-reaching fields, as evidenced in the literature \cite{DONUT}.  These include brain activity monitoring \cite{Ringach,Bio2,Bio1}, protein structural analysis \cite{Protein1,Protein2}, image processing \cite{carlsson_local_2008,Image1,Image2,gestalt}, speech recognition \cite{brown_nonlinear_2009}, audio identification \cite{ximena}, signal processing \cite{barbarossa2020topological,tulchinskii2023topological}, time series forecasting \cite{TS2} and classification \cite{TS1}, various aspects of neural networks and deep learning \cite{DL3,DL1,DL2,chen2020measuring}, among others.  

The task of extracting features that pertain to structural information is both intriguing and formidable.  This process is integral to a multitude of practical applications \cite{Sha4,Sha2,Sha1,Sha3}, as researchers strive to identify the most effective descriptors of shape within a given dataset.  Despite the fact that TDA is specifically designed for shape capture, there are several hurdles that persist in its theory and application: the nature and sensitivity of descriptors obtained by methods in TDA; dimensionality of the data and parameter choices; vectorisation of topological features for ML purposes; and computational cost.  These challenges will be elaborated in the following paragraphs within this section.  Subsequently, we will demonstrate how our proposed methodology, TopCap and TopNN, addresses these challenges through applications to speech signal processing.  

When applying TDA, the most imminent question is to comprehend the characteristics and nature of descriptors extracted via topological methods.  TDA is theoretically based in the mathematical field of algebraic topology \cite{baez,hatcher_algebraic_2002,Lee_AT}, with persistent homology (PH) being its primary tool \cite{PH1,PH2}.  While algebraic topology can quantify topological information to a certain extent, it is vitally important to understand both the capabilities and limitations of TDA \cite{carlsson_topology_2009,introDS}.  Generally speaking, TDA methods distinguish objects up to continuous deformation.  For example, PH cannot differentiate a disk from a filled rectangle, given that one can continuously deform the rectangle into a disk by pulling out its four edges.  In contrast, PH can distinguish between a filled rectangle and an unfilled one due to the presence of a hole in the latter, which prevents a continuous deformation between the two.  

In certain circumstances, these methods are considered excessively ambiguous to capture the structural information in data, and more precise descriptors of shapes are desirable.  To draw an analogy, TDA can be conceptualised as a scanner with diverse inputs encompassing time series, graphs, pictures, videos, etc.  The output of this scanner is a multiset of intervals in the extended real line, referred to as a persistence diagram (PD) or a persistence barcode (PB) \cite{persistent_shape,Barcode3,Barcode2} (cf.~Fig.~\ref{fig:ph}f and see Supplementary Sec.~\ref{sec:ph} for details, including the usual birth-by-death PDs and their birth-by-lifetime variants).  In particular, by {\em maximal persistence} (MP) we mean the maximal length of the intervals.  

\begin{figure*}
    \centering
    \includegraphics[scale=.95]{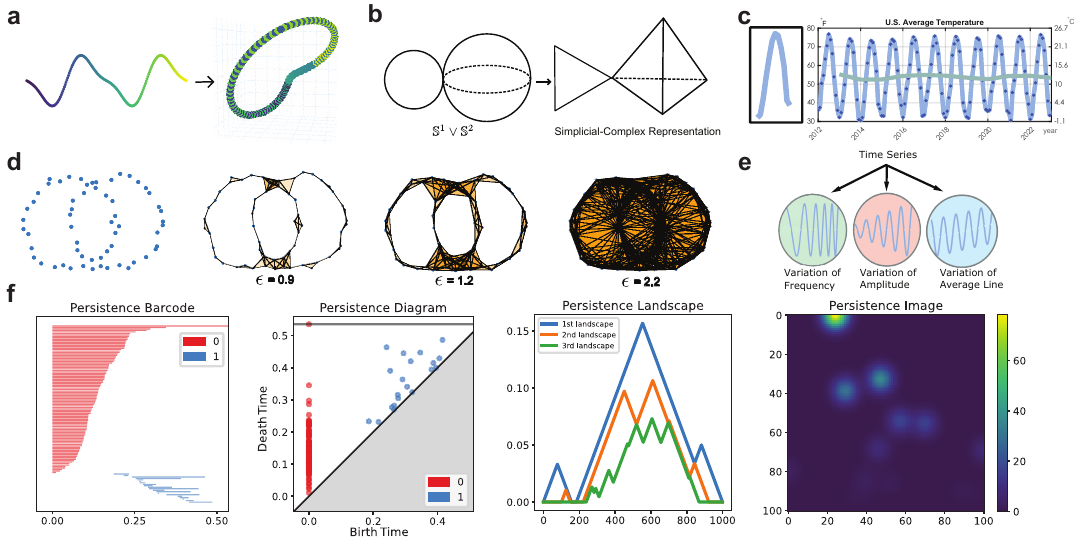}
    \caption{
        \textbf{Visualisation for data and some of their commonly used topological descriptors.}  
        \textbf{a} Time-delay embedding (dimension=3, delay=10, skip=1) of $f(t_n)=\sin(2t_n)-3\sin(t_n)$, with $t_n=\frac{\pi}{50}n$ ($0\leq n\leq200$).  Resulting point clouds lay on a closed curve in 3-dimensional Euclidean space.  The colour indicates original locations of data in the time series.  
        \textbf{b} A topological space and its triangulation.  On the left is a topological space consisting of a 1-dimensional sphere (i.e., a circle) and a 2-dimensional sphere with a single point of contact, denoted as $\mathbb{S}^1 \vee \mathbb{S}^2$.  The right depicts a triangulation of this topological space.  
        \textbf{c} Average temperature in the U.S.~with monthly values (dark blue dots) and yearly values (green curve).  The left panel shows a single-year section of average temperature.  
        \textbf{d} Computing PH.  The four plots consecutively show how a persistence diagram or barcode is computed: Connect each pair of points with a distance less than $\epsilon$ by a line segment, fill in each triple of points with mutual distances less than $\epsilon$ with a triangular region, etc., and compute the corresponding homology groups.  
        \textbf{e} Characterising the vibration of a time series in terms of its variability of frequency, amplitude, and average line.  
        \textbf{f} Commonly used representations for PH, with an example of 100 points uniformly distributed over a bounded region in 2D Euclidean space.  A persistence barcode is a multiset of intervals, where the horizontal axis shows when each feature appears and disappears.  A persistence diagram directly plots the birth and death times (values of $\epsilon$, not to be confused with time series) of each interval.  In both plots, 0 and 1 correspond to the 0-dimensional loops (connected components) and 1-dimensional loops.  In a persistence landscape, the $k$\textsuperscript{th} landscape is the $k$\textsuperscript{th} largest value of tent functions for each feature, here taken as the 1-dimensional loops, with the horizontal axis representing resolution (turning the persistence diagram clockwise by $45^\circ$).  Similarly, a persistence image is created by applying Gaussian functions centred at each feature and then converting them into a pixelated image, where both the horizontal and vertical axes represent resolution.  
    }
    \label{fig:ph}
\end{figure*}

The precision of the topological descriptor depends on two factors: (1) the association of a topological space, i.e., the process of transforming the input data into a topological space (see Fig.~\ref{fig:ph}b for a simplicial-complex representation of spaces; typically, the original datasets are less structured, and one should find a suitable representation of the data), and (2) the vectorisation of PD or PB, i.e., how to perform statistical inference with PD/PB.  Despite there are many theoretical results which provide a solid foundation for TDA, few can elucidate the practical implications of PD and PB.  For example, what does it mean if many points are distributed near the birth--death diagonal line in a PD?  Extensive studies have been conducted on short-lived bars in PH, including those related to molecular data \cite{shortBar5,shortBar3}, hierarchical structures \cite{shortBar1}, and protein structures \cite{shortBar3,shortBar2,shortBar4}, among others.  The significance of points distributed near the birth--death diagonal line is particularly relevant in real-world applications, and we shall explore it here in context as well.  

The next main challenge, as many researchers may encounter when applying topological methods, is to determine the dimension of point clouds derived from input data \cite{dimension3,dimension2,dimension1}.  This essentially involves transforming the input into a topological space.  In situations where the dimension of the data is high, researchers often project the data into a lower-dimensional topological space to facilitate visualisation and reduce computational cost \cite{Bio2,Bio1,Bio3}.  On the other hand, as in this study and other applications with time series analysis \cite{brown_nonlinear_2009,TS1,SP1,TS4,ML4_2,TS5,TS6}, low-dimensional data (on a hypothesised manifold) are embedded into a higher-dimensional (Euclidean) space.  In both scenarios, deciding on the data dimensionality is both critical and challenging.  

Often, tuning the dimension is a tremendous task.  In our case, as it might seem counterintuitive compared to most algorithms, when the data are embedded into a higher-dimensional space, the computation will be faster, the point cloud appears smoother and more regular, and most importantly, more salient topological features can be spotted, which seldom happens in lower-dimensional spaces.  When encountering the dimensionality of data, researchers would think of the well-known curse of dimensionality \cite{bellman1966dynamic}: As a typical algorithm grapple, with the increase of dimension, more data are needed to be involved, often growing exponentially and escalating computational cost.  Even worse, the computational cost of the algorithm itself normally rises as the dimension goes higher.  

However, topological methods do not necessarily prefer data of lower dimension.  For computing PH (see Fig.~\ref{fig:ph}d for the process of computing PD/PB from point clouds), a commonly used algorithm \cite{PH1,computing2} sees complexity grow with an increase in the number $n$ of simplices during the process, with a worst-case polynomial time-complexity of $O(n^3)$.  As such, the computational cost is directly related to the number of simplices formed during filtration.  Our experiments show that computation time may not increase much given an increase of dimension of data, because the latter may have little effect on the size (i.e., number of points) of the point cloud and thus neither on the number of simplices formed during filtration.  

Having obtained a suitable topological space from input data, one can derive a PD/PB from the topological space, which constitutes a multiset of intervals.  The subsequent challenge lies in the vectorisation of the PD/PB for its integration into an ML algorithm.  The vectorisation process is essentially linked to the construction of the topological space, as the combination of different methods for constructing the topological space and vectorisation together determine the descriptor utilised in ML.  A plethora of vectorisation methods exist, such as persistence landscape \cite{Vec3}, persistence image \cite{Vec4}, persistent entropy \cite{VecFun1}, and persistence curve \cite{VecFun2}, among others, as documented in various studies \cite{PH2,Vec2} (cf.~Fig.~\ref{fig:ph}f).  The selection of these methods requires careful consideration.  Additionally, one can design customised quantification techniques tailored to experimental conditions and physical properties to meet specific requirements \cite{VecFunExample1,VecFunExample3,VecFunExample2}.  
\\

\noindent
To place our study in a more specific context, let us now give an overview of closely related work in the field.  

Time series analysis \cite{hamilton2020time} is a prevalent tool for various applied sciences.  The recent surge in TDA has opened avenues for the integration of topological methods into time series analysis \cite{TS2,TS7,TS3}.  Much work in the literature has contributed to the theoretical foundation in this area.  For example, theoretical frameworks for processing periodic time series have been proposed by Perea and Harer \cite{perea_sliding_2015}, followed by their and their collaborators' implementation in discovering periodicity in gene expressions \cite{perea_sw1pers_2015}.  In \cite{perea_sliding_2015}, they studied the geometric structure of truncated Fourier series of a periodic function and its dependence on parameters in time-delay embedding (TDE), providing a solid groundwork for TopCap.  In addition to periodic time series, towards more general and complex scenarios, quasi-periodic time series have also been the subject of scholarly attention.  Research in this direction has primarily concentrated on the selection of parameters for geometric space reconstruction \cite{tralie_quasiperiodicity_2017} and extended to vector-valued time series \cite{gakhar_sliding_2021}.  

Here, a topological space is constructed from data using TDE, a technique that has been widely employed in the reconstruction of time series (see Fig.~\ref{fig:ph}a and Supplementary Sec.~\ref{sec:tde} for details).  Thanks to the topological invariance of TDE, the general construction of simplicial-complex representation (see Fig.~\ref{fig:ph}b) and computation of PH from point clouds (Fig.~\ref{fig:ph}d) both apply to time series data, although this transformation involves subtle technical issues in practice.  

For instance, Emrani et al.~utilised TDE and PH to identify the periodic structure of dynamical systems, with applications to wheeze detection in pulmonology \cite{SP1}.  They selected the embedding dimension $d$ as 2, and their delay parameter $\tau$ was determined by an autocorrelation-like (ACL) function, which provided a range for the delay between the first and second critical points of the ACL function.  

Pereira and de Mello proposed a data clustering approach based on PD \cite{TS4}.  The data were initially reconstructed by TDE, with $d=2$ and $\tau=3$, so as to obtain the corresponding PD, which was then subjected to $k$-means clustering.  The delay $\tau$ was determined using the first minimum of an auto mutual information, and the embedding dimension $d$ was set to be 2 as using 3 dimensions did not significantly improve the results.  

Khasawneh and Munch introduced a topological approach for examining the stability of a class of nonlinear stochastic delay equations \cite{ML4_2}.  They used false nearest neighbours to determine the embedding dimension $d=3$ and chose the delay to equal the first zeros of the ACL function.  Subsequently, the longest persistence lifetime in PD was used for vectorisation to quantify periodicity.  

Umeda focused on a classification problem for volatile time series by extracting the structure of attractors, using TDA to represent transition rules of the time series \cite{TS1}.  He assigned $d=3$, $\tau=1$ in his study and introduced a vectorisation method, which was then applied to a convolutional neural network (CNN) to achieve classification.  

Gidea and Katz employed TDA to detect early signs prior to financial crashes \cite{TS6}.  They studied multi-dimensional time series with $\tau=1$ and used persistence landscape as a vectorisation method.  

For speech recognition, Brown and Knudson examined the structure of point clouds obtained via TDE of human speech signals \cite{brown_nonlinear_2009}.  The TDE parameters were set as $d=3$, $\tau=20$, based on which they examined the structure of point clouds and their corresponding PBs.  

Upon reviewing the relevant literature, we see that currently there is still lack of a general framework for systematically choosing $d$ and $\tau$ with machine learning, and researchers often have to make choices in an ad hoc fashion for practical needs.  
\\

\noindent
In this work, motivated by and aiming at real-world applications to artificial intelligence, we develop methods for topological speech (and audio) signal processing, beyond direct biomimetic spectral engineering currently adopted in the field (cf.~Supplementary Sec.~\ref{sec:phone}).  The first main method, TopCap, is a streamlined combination of topological data analysis to machine learning, with fine-tuned time-delay embedding juxtaposed with persistent homology on the topological end, followed by accessible, user-friendly machine learning algorithms.  The second main method, TopNN, provides state-of-the-art neural networks for audio and speech signal processing, such as gated recurrent units, with topology enhancement by concatenating black-box neural network feature vectors with interpretable TopCap feature vectors for decoder.  These methods extract and integrate topological features of phonetic data beyond those obtained via short-time Fourier transform or mel-frequency cepstral coefficients.  

Applying TopCap and TopNN to the task of classifying voiced and voiceless consonants, we obtain the following main results.  Firstly, in terms of accuracy, TopCap stands in comparison with various state-of-the-art models across a wide range of small and large datasets.  In addition, it shows advantages in structural efficiency, interpretability, and computational cost.  Secondly, compared to state-of-the-art neural networks, our experiments with TopNN demonstrate better accuracy, steadier performance, and more robustness against noise.  

Besides, for experts working on topological time series analysis and on nonlinear time series analysis, we offer the following conclusions.  Firstly, noisy or complex real-world time series require a parameter selection scheme for time-delay embedding that goes beyond the Perea--Harer framework.  Notably, the significant topological feature of maximal persistence exhibits extreme sensitivity to the delay parameter, while it correlates sublinearly to the embedding dimension.  This latter finding of higher embedding dimension for more prominent topological features (and consequently better overall performance), seemingly paradoxical, stands in stark contrast to the common intuition from the curse of dimensionality as well as to the relatively low intrinsic dimensions of time series data.  Secondly, preliminary experiments with both synthetic and real-world data show the capability and potential of topological representations, such as persistence diagrams (utilising points distributed near the birth--death diagonal line), in capturing and distinguishing finer patterns of vibration that go beyond periodicity, namely, variation of frequency, of amplitude, and of average line.  Thirdly, we propose formant spectral features and cyclic time-delay embedding configuration eigenvalues as additional geometric features for phoneme recognition.  

Our research drew inspiration from Carlsson and his collaborators' discovery of the Klein-bottle distribution of high-contrast, local patches of natural images \cite{carlsson_local_2008}, as well as their subsequent recent work on topological CNNs for learning image and even video data \cite{Gabrielsson2019,ML2,DL3}.  By analogy, based on our first findings in this direction, we aim to understand a distribution space for speech data, even a directed graph structure on it modelling the complex network of speech-signal sequences for practical purposes such as speaker diarisation.  Moreover, we aim to better understand how these topological inputs enable smarter learning in relation to human auditory perception (cf.~\cite{Ringach}).

\section*{Results}

\begin{figure*}
    \centering
    \includegraphics{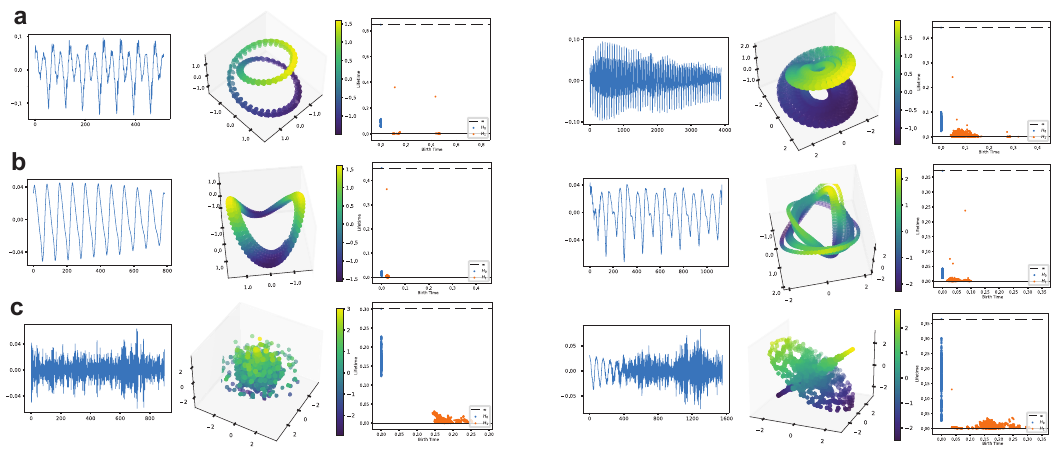}
    \caption{
        \textbf{The varied shapes of vowels, voiced consonants, and voiceless consonants.}  
        \textbf{a} The left three panels and the right three panels depict two vowels, respectively.  For each triple, the first picture is the waveform of the vowel, the second picture corresponds to the 3-dimensional principal component analysis of the point cloud resulting from performing TDE (dimension=100, delay=1, skip=1) on this time series (the colour legend shows the vertical coordinate), and the third picture is the PD of this point cloud.  
        \textbf{b} The analogous features for two voiced consonants.  
        \textbf{c} Those for two voiceless consonants.  
    }
    \label{fig:shape}
\end{figure*}

In this section, we present our experiments and results using methods for topological speech signal processing.  Let us briefly summarise them as follows.  

Firstly, we propose TopCap, a framework that embeds speech signals into high-dimensional space using time-delay embedding, and then extracts significant topological features via persistent homology.  Such features serve as representations of a signal’s periodicity.  Their topological descriptors are subsequently fed into traditional machine learning algorithms for classification.  

In particular, we benchmark TopCap against several state-of-the-art models based on neural networks for speech processing.  The results show that TopCap achieves comparable classification accuracy while offering improved efficiency and interpretability.  

To further compare the feature extraction approach of TopCap with traditional speech signal processing methods, we conduct a feature analysis.  The results demonstrate that the features extracted by TopCap exhibit stronger discriminative power, rendering them more effective for consonant classification.  

Secondly, motivated by the complementary strengths of topological and deep learning approaches, we propose TopNN, namely, topology-enhanced neural networks.  In this model, the encoder integrates topological features with features extracted by a neural network, which are then passed through a decoder composed of fully connected layers to map the representation to the target labels.  Experimental results demonstrate that TopNN successfully combines the advantages of both paradigms, yielding significant improvements in classification accuracy, robustness to noise, and model stability.  

Thirdly, beyond the experiments focused on capturing periodicity in time series data, we conduct preliminary studies using both synthetic and real-world datasets to explore the broader potential of topological representations.  Our findings suggest that persistent diagrams—particularly through the analysis of points near the birth--death diagonal—can effectively capture and distinguish more nuanced vibration patterns beyond periodicity, including variations of frequency, amplitude, and average line.  

As a snapshot for the objects of study and their multifaceted profiles leading to our findings, Fig.~\ref{fig:shape} gives an intuitive visualisation for vowels, voiced consonants, and voiceless consonants in TDE and PD, respectively (see Supplementary Sec.~\ref{sec:phone} for details of phonetic categories).

\subsection*{TopCap: topological features for machine learning}

In this subsection, we present our results on consonant recognition using topology-enhanced machine learning methods, notably, the streamlined approach of TopCap.  The classification of voiced and voiceless consonants serves as a significant, relevant application of our methodology, showcasing its efficacy and advantages.  Meanwhile, as a hands-on example originating directly from industrial innovation, it makes various technical considerations in developing our methods more transparent and highlights potential for further investigation and enhancement.  

Voiced and voiceless regions of speech have distinct speech production processes and energy patterns.  Segmentation of voiced and voiceless speech is a fundamental and important process for various speech processing applications \cite{rabiner1993fundamentals}.  In medical diagnosis, researchers can detect common cold and other diseases by studying voiceless and voiced sounds \cite{Kao,Warule}.  The detection of voiced and voiceless sounds can also be used to reveal whether musical expertise leads to an altered neurophysiological processing of sub-segmental information available in the speech signal \cite{Ott}.  It is particularly important to study the segmentation of voiced and voiceless sounds in linguistics, and a variety of methods have been developed and applied \cite{Bancroft,Huiqun,Bachu,Zhaoting}.  Moreover, there are applications geared towards AI innovations, for example, speaker identification via voiceless consonants \cite{Juan}.  Thus, it has become imperative to research the characteristics of voiced and voiceless sounds and distinguish them, which can ensure the accuracy of the segmentation and enable other applications.  Placed in a broader context, this analysis for speech recognition at the phonemic level precedes the type of higher-order language processing typically associated with NLP (cf.~\cite{whale}).  

Given consonant recognition as a significant problem originating and posed to us from the industry, we perform multiple topology-enhanced machine learning experiments as follows.  \\

\begin{figure*}
    \centering
    \includegraphics{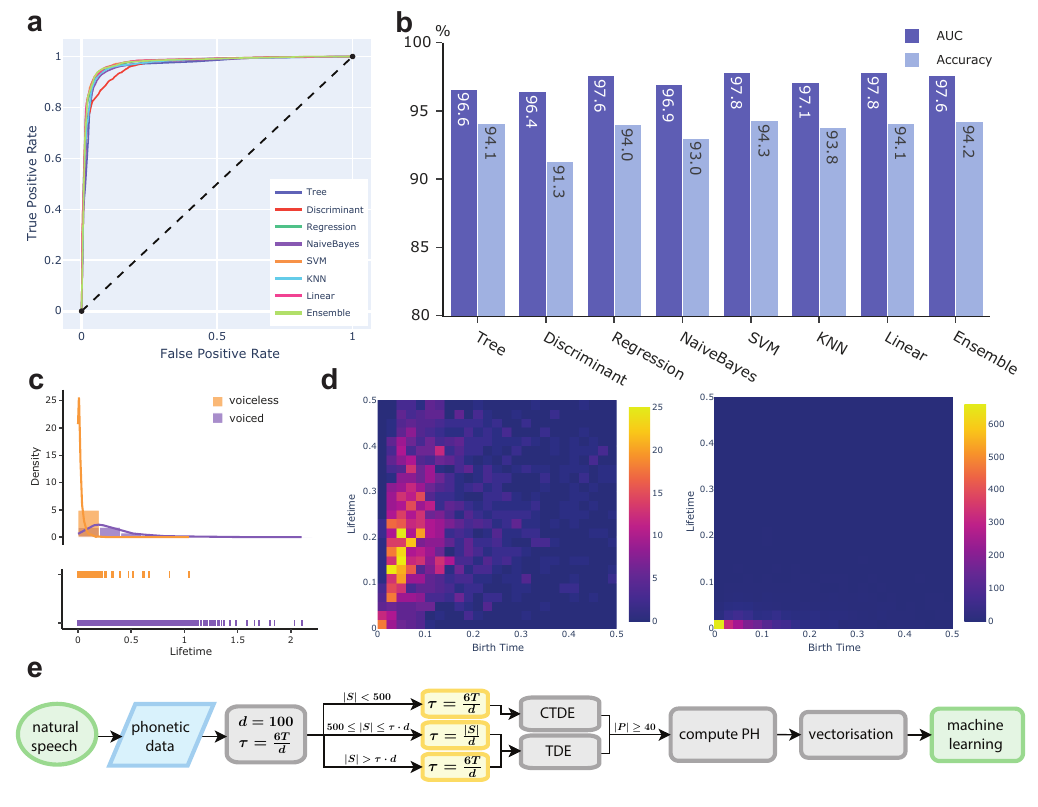}
    \caption{
        \textbf{Results of machine learning topological features.}  
        \textbf{a} ROCs of traditional ML algorithms.  
        \textbf{b} Accuracy and AUC of each of these algorithms.  
        \textbf{c} Histograms of records represented by their PH-lifetime for voiced and voiceless consonants, together with kernel density estimation and rug plot.  The distributions of MP can distinguish voiced and voiceless consonants.  
        \textbf{d} Diagrams of records represented as (birth time, lifetime) for voiced consonants (left) and voiceless consonants (right), where voiced consonants exhibit higher birth time and lifetime.  The colour represents the density of points in each unit grid box.  The features (birth time, lifetime) interpret the most prominent structural feature and its PH-birth time.  
        \textbf{e} Flowchart of the primary experiments.  Here $|S|$ denotes the number of samples in a time series, $|P|$ denotes the number of points in the point cloud, and $T$ denotes the (minimal) period of the time series computed by the ACL function.  
    }
    \label{fig:topcap}
\end{figure*}

\noindent
Using datasets comprising human speech, we first employ the Montreal Forced Aligner (MFA) \cite{mcauliffe17_interspeech} to align natural speech signals into phonetic segments.  Following preprocessing of these phonetic segments, TDE is conducted with dimension parameter $d=100$ and delay parameter $\tau=6T/d$, where $T$ approximates the (minimal) period of the time series.  Following additional refinement procedures, PDs are computed for these segments and are then vectorised based on MP and its corresponding birth time.  The comprehensive procedural framework is presented in more detail in Supplementary Sec.~\ref{sec:method}, while the corresponding workflow is shown in Fig.~\ref{fig:topcap}e.  

It is worth noting that in the applications of TDE, the dimension parameter $d$ is usually determined through some algorithms designed to identify the minimal appropriate dimension \cite{dimension3,kantz_schreiber_2003}.  Here, the embedding dimension $d=100$ was chosen to be as large as possible within the constraints of our data.  More specifically, in our experiments, using lower dimensions such as $d=5$, 10, or 20 yielded poor results, as those dimensions were insufficient to capture the complex underlying structure of the time series.  In higher dimensions, important features that are not apparent in lower dimensions become much easier to identify.  However, the dimension cannot be too large either, otherwise the embedded point cloud obtained following the theoretical framework of Perea and Harer \cite{perea_sliding_2015} may consist of too few points to adequately represent the original data structure (see Supplementary Sec.~\ref{subsec:m_tde} for details).  The delay parameter $\tau$ is determined by an ACL function with no specific rule, but in many cases $\tau=mT/d$ for some positive integer $m$.  In our pursuit of enhanced extraction of topological features, a relatively high dimension is chosen (see Discussion for more discussion on dimension in TDE).  Given this higher dimension, the usual case of $\tau=T/d$ with $m=1$ may prove excessively diminutive, particularly in light of the time series only taking values in discrete time steps.  Consequently, in TopCap we adopt an adjusted parametrisation for $\tau=mT/d$ with a relatively large value $m=6$.  

We input the pair of MP and birth time from 1-dimensional PD for each sound record to multiple traditional classification algorithms: Tree, Discriminant, Regression, Naive Bayes, Support Vector Machine, $k$-Nearest Neighbours, Linear, and Ensemble.  We use the application of the MATLAB (R2024b) Classification Learner, with 5-fold cross-validation, and set aside 30\% records as test data.  This application performs machine learning algorithms in an automatic way.  There are a total of 5101 records, with 3571 training samples and 1530 test samples.  Among them, 2138 records are voiced consonants and the remaining 2963 records are voiceless consonants.  The models we choose in this application are Optimizable Tree, Optimizable Discriminant, Binary GLM Logistic Regression, Optimizable Naive Bayes, Optimizable SVM, Optimizable KNN, Optimizable Efficient Linear, and Optimizable Ensemble.  In this case study, we employ traditional machine learning techniques to address the classification problem, using relatively small datasets to illustrate performance.  In subsequent cases involving deep learning algorithms, substantially larger datasets will be used, as discussed in the following section.  

The results are shown in Fig.~\ref{fig:topcap}a--d.  The receiver operating characteristic curve (ROC), area under the curve (AUC), and accuracy metrics collectively demonstrate the efficacy of these topological features as inputs for a variety of machine learning algorithms.  Most of the algorithms incorporating topological inputs attain AUC and accuracy surpassing 96\% and 93\%, respectively.  The ROC and AUC together depict the high performance of our classification model across all classification thresholds.  The 2D histograms depicted in Fig.~\ref{fig:topcap}c--d collectively illustrate the distinct distributions of voiced and voiceless consonants: Voiced consonants tend to exhibit a higher birth time and lifetime.  As an explanation for the high performance of these algorithms, while a persistence diagram may consist of many points and is hard to vectorise, the birth time and lifetime corresponding to the most persistent feature are sufficient for the classification of voiced and voiceless consonants.  Despite the intricate structure that a PD may present, appropriately extracted topological features enable traditional machine learning algorithms to separate complex data effectively.  This highlights the potential of TDA in enhancing the performance of machine learning models.  \\

\noindent
We next demonstrate the advantages of TopCap by comparing it with state-of-the-art methods in speech recognition that are not based on topology, over a diverse range of benchmark datasets.  

In the above primary experiments, our analysis solely utilised the HT1 corpus sourced from the broader ALLSSTAR dataset of SpeechBox \cite{SpeechBox} (see Supplementary Sec.~\ref{subsec:m_pre} for details).  We extend this by conducting a series of experiments across a diverse array of datasets using the same methodology, with the aim of enhancing the robustness and credibility of our results.  These datasets encompass renowned benchmark repositories such as LJSpeech \cite{LJSpeech}, TIMIT \cite{TIMIT}, and LibriSpeech \cite{LibriSpeech}, in addition to supplementary corpora sourced from ALLSSTAR.  Collectively, they contain a substantial amount of phones, numbering in the hundreds of thousands: LJSpeech provides around 200000, TIMIT around 40000, LibriSpeech over 7000000 (1000 hours of speech), and ALLSSTAR around 20000 in total.  

For comparative analysis with existing methodologies, we have placed our approach alongside three methods that are not based on topology.  Specifically, we combine standard audio processing methods for feature extraction with state-of-the-art deep learning methods for classification tasks.  The former methods include short-time Fourier transform (STFT) and mel-frequency cepstral coefficients (MFCC).  The latter methods include CNNs,  gated recurrent units (GRU) networks, and Transformers.  As such, we perform experiments on the above datasets using the methods of STFT--CNN, MFCC--GRU, and MFCC--Transformer, in comparison with those using TopCap.  

In more detail, TopCap comprises TDE--PH and an array of traditional, accessible machine learning methods.  The coupling of TDE and PH serves to extract the latent topological features within the time series, while STFT and MFCC each extract features through analytic methods based on spectral information.  Our selection of the multiple machine learning and deep learning architectures in each experimental pipeline is informed by the nature of the extracted features. Specifically, the output spectrograms from STFT are imagery representations, making them well suited for CNNs.  In particular, we design and compare two models for this method, denoted by STFT--CNN-8 and STFT--CNN-16.  The former resizes each grey-scale spectrogram of 124$\times$129 pixels through bilinear interpolation down to 8$\times$8 with 386177 parameters, while the latter to 16$\times$16 with 435329 parameters (a 90\% reduction of parameters from the original 124$\times$129 neural network).  Both networks consist of 5 layers with 3 convolutional and 2 fully connected.  In contrast, MFCC features, characterised by their lower dimensionality, are more appropriate for recurrent-neural-network architectures, such as GRUs and Transformers.  

\begin{table*}
    \centering
    \begin{NiceTabular}{l|P{.8cm}P{.8cm}P{.8cm}P{.8cm}P{.8cm}|P{.8cm}P{.8cm}P{.8cm}}
        \toprule
        & & \multicolumn{3}{c}{ALLSSTAR corpora} & & \multicolumn{3}{c}{Random samples} \\
        \midrule
        Small dataset & HT1 & HT2 & DHR & LPP & NWS & LJ & TIMIT & Libri \\
        Number of phones & 3200 & 3000 & 3600 & 3800 & 1800 & 2000 & 2000 & 2000 \\
        \RowStyle{\bfseries} TopCap & 94.3 & 92.7 & 92.3 & 91.9 & 88.8 & 94.6 & 83.9 & 85.1 \\
        MFCC--GRU                   & 93.3 & 92.2 & 93.2 & 91.4 & 89.8 & 86.0 & 70.5 & 79.0 \\
        MFCC--Transformer           & 96.0 & 93.9 & 94.2 & 92.4 & 94.4 & 92.0 & 96.3 & 87.5 \\
        STFT--CNN-8                 & 87.1 & 84.0 & 78.2 & 79.1 & 79.9 & 82.7 & 76.3 & 77.5 \\
        STFT--CNN-16                & 96.7 & 95.1 & 94.4 & 92.1 & 94.0 & 95.6 & 89.4 & 88.7 \\
        \midrule[1pt]
        Large dataset & \multicolumn{2}{c}{ALLSSTAR} & \multicolumn{2}{c}{LJSpeech} & \multicolumn{2}{c}{TIMIT} & \multicolumn{2}{c}{LibriSpeech} \\
        Number of phones & \multicolumn{2}{c}{21000} & \multicolumn{2}{c}{257000} & \multicolumn{2}{c}{42000} & \multicolumn{2}{c}{500000} \\
        \RowStyle{\bfseries} TopCap & \multicolumn{2}{c}{92.5} & \multicolumn{2}{c}{92.9} & \multicolumn{2}{c}{92.8} & \multicolumn{2}{c}{88.7} \\
        MFCC--GRU                   & \multicolumn{2}{c}{93.9} & \multicolumn{2}{c}{96.2} & \multicolumn{2}{c}{97.4} & \multicolumn{2}{c}{91.0} \\
        MFCC--Transformer           & \multicolumn{2}{c}{93.7} & \multicolumn{2}{c}{96.9} & \multicolumn{2}{c}{97.6} & \multicolumn{2}{c}{92.1} \\
        STFT--CNN-8                 & \multicolumn{2}{c}{81.2} & \multicolumn{2}{c}{85.4} & \multicolumn{2}{c}{77.5} & \multicolumn{2}{c}{80.3} \\
        STFT--CNN-16                & \multicolumn{2}{c}{94.6} & \multicolumn{2}{c}{96.3} & \multicolumn{2}{c}{91.4} & \multicolumn{2}{c}{90.6} \\
        \bottomrule
    \end{NiceTabular}
    \caption{
        \textbf{Accuracy rates \% of TopCap (in bold) on 8 small datasets and 4 large datasets stand in comparison with state-of-the-art methods.}  
        The random samples are taken from the large datasets listed in the lower half of the table.  In particular, in the second row, LJ and Libri are abbreviations for LJSpeech and LibriSpeech, respectively.  While MFCC--Transformer and STFT--CNN-16 generally outperform TopCap, it is important to note that TopCap exceeds the performance of MFCC--GRU (which also uses advanced architecture) and STFT--CNN-8 (a smaller model than STFT--CNN-16) on small datasets.  For larger datasets, TopCap generally does not match the performance of deep neural networks, primarily due to its use of simpler topological features and basic machine learning models.  This limitation motivates the integration of topological features into neural networks, as discussed below.  Overall, while TopCap may not achieve the highest performance across all benchmarks, it produces decent results.  
    }
    \label{tab:topcap}
\end{table*}

Table \ref{tab:topcap} presents the results of our experiments with TopCap and the comparative models on benchmark datasets listed above.  In each of the upper and lower halves, on the top row, the various datasets are displayed.  The remaining rows record the data sizes (i.e., numbers of phones) along with the corresponding accuracy rates of TopCap and of the comparative models applied to these datasets.  In the upper half of Table \ref{tab:topcap}, we focus on small-scale datasets.  The five subsets of ALLSSTAR each comprise their entire phones, while LJSpeech, TIMIT, and LibriSpeech datasets are sampled randomly, each containing 2000 samples with 50\% voiced consonants and 50\% voiceless consonants.  The lower half of Table \ref{tab:topcap} displays the results from large-scale datasets.  Among them, ALLSSTAR, LJSpeech, and TIMIT each contribute their entire data for analysis, while LibriSpeech contributes 500000 phones out of 1800000 from its speech data (we obtained 1800000 phonetic segments from a half of the 500-hour speech data).  A main consideration for dividing the experiments into small and large datasets lies in the nature of training and generalisation for neural networks, which depend on the size of a dataset and correlate with the networks' performances.  

These results show that, in classification of voiced and voiceless consonants, our topology-enhanced model TopCap achieved competent accuracy rates on small datasets and sustained a decent performance on larger ones, in comparison with state-of-the-art models that are not based on topology.  Moreover, our topology-enhanced approach shows significant advantages in the following three areas.  

The first area is structural efficiency.  Neural network models require further feature extraction from input MFCC sequences or STFT spectrograms for classification tasks, necessitating a training process which lengthens with a growing dataset.  In contrast, TopCap mainly utilises topology-based methods (TDE and PH) which are more straightforward for feature extraction.  Meanwhile, the topological fingerprints (e.g., maximal persistence) are strong enough to characterise phonemes effectively for our classification tasks (see also the feature analysis below).  Therefore, TopCap gains higher efficiency, especially when handling larger datasets.  On a related note, deep learning methods, as a data-driven approach, require large amounts of data for training and generalisation.  In contrast, comparing the upper and lower halves of Table \ref{tab:topcap}, we see that TopCap achieves equally good performance on relatively small datasets.  

The second area is interpretability.  Neural networks are often referred to as black boxes due to their low explainability and interpretability, which makes it challenging to understand the mechanism of feature extraction and effectively improve a model for classification.  However, TopCap offers a white-box method for visualising features of time series data, which gives insight to the intrinsic properties and nuanced differences within the data, enabling us to better understand and improve the model (see TopNN below).  

The third area is computational speed.  Neural networks involve time-consuming training process, even with GPU acceleration.  For instance, on the TIMIT dataset, a full training cycle of 15 epochs can take approximately 30 minutes with GPU parallelisation.  In contrast, TopCap bypasses the need for iterative training and achieves significantly faster computation.  TopCap performs lightweight machine learning with negligible runtime overhead, completing both feature extraction and classification in just 2 minutes when utilising 16-thread CPU parallelisation.  TopCap's efficiency advantage comes from avoiding gradient-based optimisation and using computationally cheaper topologically derived features, along with a highly parallelisable pipeline.  These make it significantly faster and more scalable especially for large datasets or real-time applications.  

To further enhance the computational efficiency of the periodicity detection module in the TopCap algorithm, we can transition from using the auto-correlation function (ACF) to the Fast Fourier Transform (FFT), a modification primarily driven by performance considerations.  While FFT's $O(N\cdot\log N)$ complexity offers a significant speed advantage over ACF's $O(N^2)$ approach, particularly beneficial for large-scale datasets, we observed nuanced accuracy variations.  This FFT-based approach, computationally comparable to MFCC extraction in neural networks, is adopted for the subsequent experiments with TopNN below.  \\

\begin{figure*}
    \centering
    \includegraphics[scale=.95]{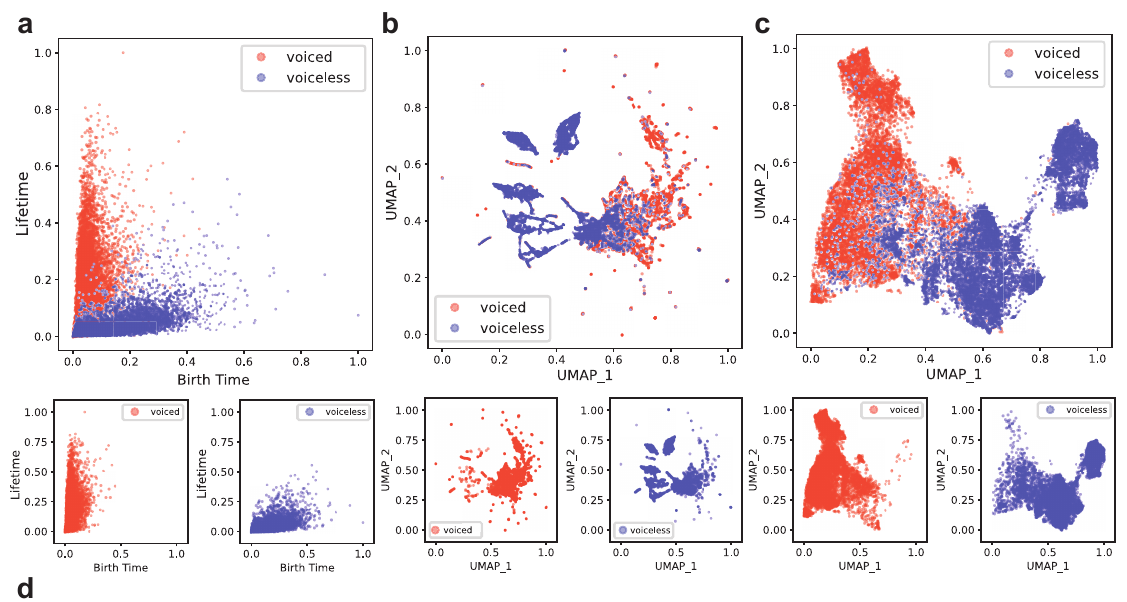}
    \textcolor{white}{.......}
    \begin{tabular}{l|P{3.8cm}|P{3.8cm}P{3.8cm}}
        \toprule
                                      & Topological features & \multicolumn{2}{c}{Spectral features} \\
        \midrule
        Method                        & TopCap               & STFT            & MFCC \\
        Accuracy \%                   & $89.66\pm0.46$       & $88.45\pm0.45$  & $88.96\pm0.42$ \\
        95\% confidence interval \%   & $[88.86,90.44]$      & $[87.64,89.43]$ & $[88.16,89.68]$ \\
        \bottomrule
    \end{tabular}
    \caption{
        \textbf{Comparative analysis of the features derived from TopCap, STFT, and MFCC.}  
        \textbf{a} Features from TopCap shown via PD.  The upper plot displays an overall view of both voiced and voiceless features, while the two lower plots provide individual representations for the voiced and voiceless categories.  Subplots (b) and (c) adhere to this layout.  Here, voiced data typically exhibit longer lifetimes with earlier birth times, whereas voiceless data tend to show shorter lifetimes with later birth times.  A small subset of both voiced and voiceless data in the middle region overlap.  
        \textbf{b} Features from STFT shown via UMAP.  Voiced data form a single cluster, whereas voiceless data are distributed across multiple clusters.  
        \textbf{c} Features from MFCC shown via UMAP.  Most voiced data group into a single cluster, with a small subset forming another cluster in the upper-left region.  Voiceless data primarily form two distinct clusters.  
        \textbf{d} A quantitative comparison via logistic regression of the separability among topological and spectral features.  The classification accuracy rates are shown as mean\,$\pm$\,standard deviation through repeated stratified cross-validation.  They demonstrate that topological features stand in comparison and even outperform traditional spectral features.  
    }
    \label{fig:feature}
\end{figure*}

\noindent
To highlight the advantages of our model in feature extraction, we conduct a feature analysis by comparing the features generated by TopCap, STFT, and MFCC.  The data utilised for this analysis is sourced from the LJSpeech dataset \cite{LJSpeech}, with a random sampling of 9000 voiced and 9000 voiceless data points from the entire library.  

For topological features, we use the same algorithm as TopCap outlined above (more details in Supplementary Sec.~\ref{sec:method}), deriving the birth time and lifetime for each sampled phone.  

For traditional spectral features, in the case of STFT, we divide each sample into time segments, perform Fourier transformation, and extract the resulting dominant frequencies as feature representation.  Here, we divide the data into six time segments in accordance with industry standards.  Specifically, given a sampling rate of 22050 Hz and an ideal speech data window length of 20--40 ms, considering that the average data length is 1654 subsamples, we set the window length to be a third of each data length, so that each window lasts approximately 25 ms.  In this way we divide each sample into six time segments of equal length, each 12.5 ms, and record the dominant frequencies of the six overlapping windows (including an extended one by the algorithm).  

In the case of MFCC, we directly apply the MFCC technique to the data, yielding 50 features that characterise the spectral properties of an audio sample.  Here, each time window returns 10 MFCCs, which give rise to 50 features in total.  As the typical number of MFCCs for speech data is 12, our choice of 10 aligns with standard practice and ensures a fair comparison.  

To visualise the spectral features, we employ uniform manifold approximation and projection (UMAP) to reduce the dimensionality to 2 dimensions, from STFT's 6 dimensions and MFCC's 50 dimensions.  Finally, we do a min-max normalisation to all the 2D features, including the PD from TopCap.  In addition, to quantitatively compare the separability among these 2D features, we apply a logistic model to classify the data based on them.  Specifically, for each feature we run stratified 5-fold cross-validation and repeat it 10 times.  We then compute the mean, standard deviation, and 95\% confidence interval of those 50 accuracy rates.  The results are presented in Fig.~\ref{fig:feature}.  

Often, feature extraction techniques such as STFT and MFCC capture information in high dimensions.  However, in order to effectively use these high-dimensional features, dimensionality reduction techniques such as UMAP and $t$-SNE ($t$-distributed stochastic neighbour embedding) are often applied to visualise data or reduce their complexity.  A primary issue here is the potential loss of structural information from the original features.  For example, in Fig.~\ref{fig:feature}b, while UMAP reduces the dimensionality to 2, the physical meaning of these 2 dimensions is unclear.  More importantly, it is difficult to discern the original data structure based on the reduced representation.  Specifically, in Fig.~\ref{fig:feature}c, there are two clusters in voiceless data, and it is unclear how these clusters correspond to the structure of the original data.  In contrast, our topological approach records the structure of a point cloud in a persistence diagram, from which we observe topological structure that is most persistent, so that we reduce the dimensionality while keeping the data structure transparent.  

A natural question then arises: If we increase the feature dimensionality for STFT and MFCC, will the separability results surpass those of TopCap?  This is likely, as higher dimensionality generally carries more nuanced information.  In TopCap, we retain only a single point corresponding to the longest lifetime, disregarding all other information from persistent homology.  While this approach preserves the original structure and performs reasonably well, it is inherently less informative.  As a stand-alone topological method, TopCap remains limited.  In the next subsection, we shall integrate TopCap with deep neural networks.  On one hand, neural networks are highly expressive and generate rich representations.  On the other hand, topological methods preserve structural information under high noise levels and recover geometric details that neural networks may overlook.

\subsection*{TopNN: topology-enhanced neural networks}

In the previous subsection, we proposed TopCap, which integrates topological methods with traditional machine learning approaches (e.g., KNN) for consonant classification, and compared it against state-of-the-art models (e.g., MFCC--GRU).  The experimental comparison above reveals that neural network models achieve excellent classification accuracy in specific scenarios owing to their complex architectures, while topological methods demonstrate significant advantages over neural networks in computational efficiency, model stability, and interpretability.  

Here, motivated by the complementarity between these two paradigms, we further develop topology-enhanced neural networks, a framework that combines topological feature extraction with neural architectures.  This hybrid model enables significant improvement in classification accuracy, noise resistance, robustness, and stability in consonant classification experiments.  \\

\begin{figure*}
    \centering
    \includegraphics[scale=.95]{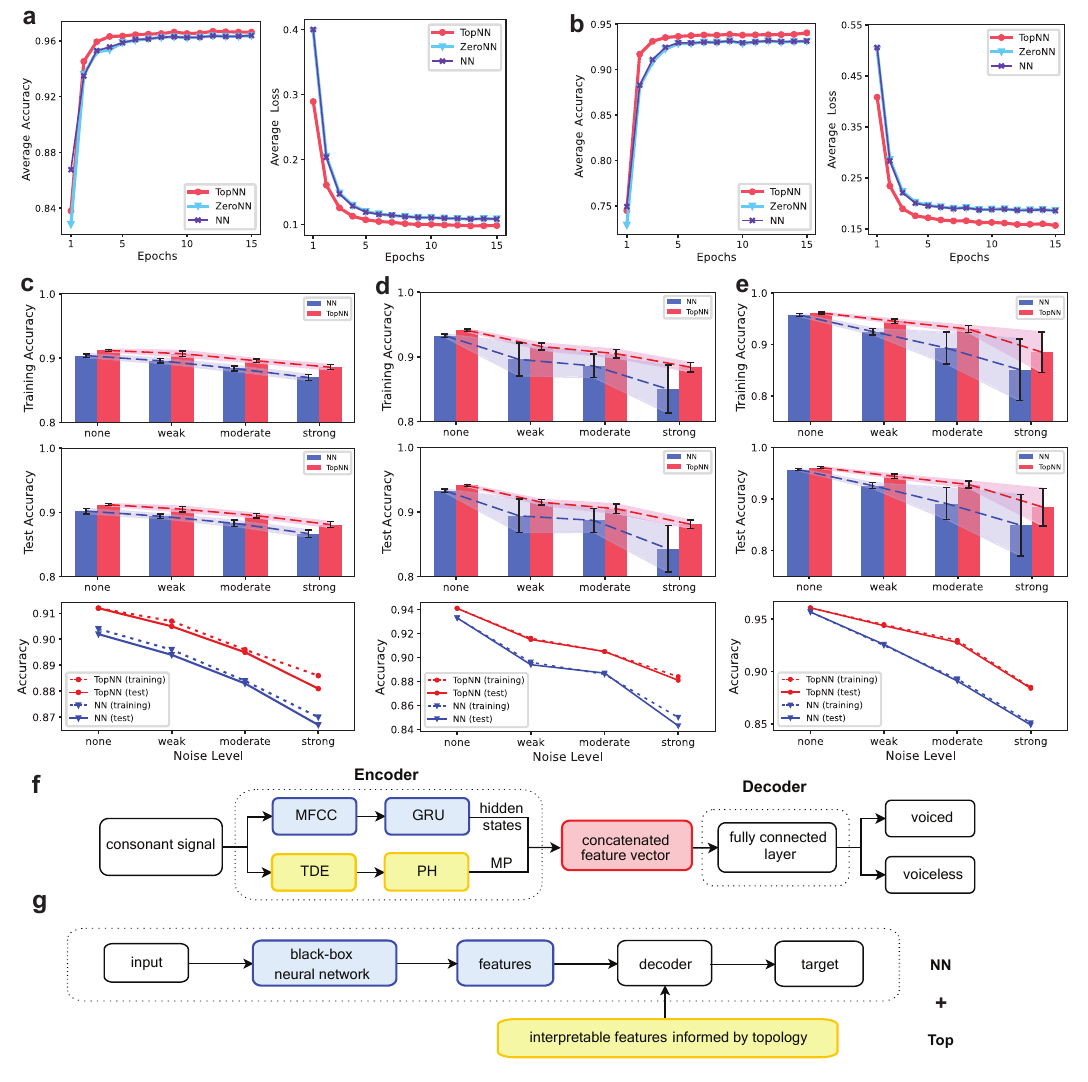}
    \caption{
        \textbf{Visual analytics of experiments with topology-enhanced neural networks TopNN.}  
        \textbf{a} Training curves of TopNN, ZeroNN (NN features concatenated with null topological feature, as a sanity check), and NN on 36000 original speech data from the TIMIT dataset.  They demonstrate that TopNN has higher accuracy and faster convergence in loss function than ZeroNN and NN.  
        \textbf{b} Training curves of TopNN, ZeroNN, and NN with the same set up as in (a) and including noise (SNR\,=\,5dB).  With Gaussian white noise added, TopNN's improvement in accuracy and loss decrease are more prominent compared with the results in (a).  
        \textbf{c}--\textbf{e} Comprehensive performance comparison and noise robustness analysis of TopNN and NN based on training and test accuracy rates with the large datasets ALLSSTAR, LJSpeech, and TIMIT from Table \ref{tab:topcap}, respectively.  Noise levels include none, weak (SNR\,=\,10dB), moderate (SNR\,=\,5dB), and strong (SNR\,=\,0dB).  In all three figures, TopNN achieves higher accuracy and is more robust against noise than NN.  
        \textbf{f} Architecture of the specific TopNN used above.  
        \textbf{g} A generic flowchart for enhancing neural networks with topological features.  
    }
    \label{fig:topnn}
\end{figure*}

\noindent
Topology-enhanced neural networks are fundamentally structured based on an encoder--decoder architecture.  The encoder comprises both black-box neural networks and topological feature extraction modules, each responsible for capturing distinct features from different aspects of the data.  These features are subsequently fused to form a comprehensive latent representation.  The decoder, usually constructed using neural networks, learns to assign optimal weights to these heterogeneous features through training and transform the features into the target variable.  Fig.~\ref{fig:topnn}g presents a conceptual framework for topology-enhanced neural networks, which enhance neural networks with interpretable features informed by topology.  

For consonant recognition, as shown more specifically in Fig.~\ref{fig:topnn}f, we propose TopNN, an integrated architecture that combines topological feature extraction modules (TDE--PH) and GRU to serve as an encoder, followed by a decoder composed of fully connected layers for classification.  Our model concatenates the MP features obtained by TopCap with the final hidden states extracted by GRU, forming a joint feature vector.  These combined representations are then processed through fully connected layers to learn the weights of different features for the voiced/voiceless classification.  \\

\noindent
In the voiced/voiceless consonant classification experiments, consonant signals are fed into TopNN for hierarchical feature extraction and classification.  

To establish a robust baseline for comparative analysis, we designate NN (with standard GRU as encoder and fully connected layer as decoder) as a baseline model, and introduce ZeroNN, an ablated variant of TopNN where topological features are replaced with zero vectors.  In NN, the encoder extracts a 6-dimensional feature vector derived from the GRU.  In contrast, the encoder in TopNN extracts a 7-dimensional feature vector by concatenating the 6-dimensional GRU-derived features with a 1-dimensional topological descriptor, i.e., the maximal persistence extracted from a PD.  In ZeroNN, the encoder outputs a 7-dimensional feature vector constructed by concatenating the GRU-extracted 6-dimensional features with an additional 1-dimensional zero.  
This controlled experimental design ensures any observed performance differences are exclusively attributable to topological feature incorporation.  

Moreover, to fully demonstrate the advantage of our topology-enhanced model, particularly its resilience and robustness derived from topological properties, we conduct comprehensive noise injection experiments on speech data across four signal-to-noise ratio (SNR) levels: the original data (SNR\,=\,$\infty$), weak noise (SNR\,=\,10dB), moderate noise (SNR\,=\,5dB), and strong noise (SNR\,=\,0dB) conditions.  The injected white noise follows a Gaussian amplitude distribution, carefully selected to emulate the natural characteristics of electronic device background noise and independent of the speech signals.  This provides a realistic simulation of real-world acoustic interference and sticks to industrial practice.  

We systematically evaluate the classification performance of TopNN, ZeroNN, and the standard NN on both the original and the noise-added speech data.  Fig.~\ref{fig:topnn}a--b track the three models' training progression on original and noise-added speech data, respectively.  

To reduce performance fluctuation arising from data selection bias and enhance the reliability of our comparisons, we employ a 5-fold cross-validation strategy.  In each fold, training and test data are randomly sampled, allowing for a more comprehensive assessment of model generalisation.  We conduct multiple experiments and use the mean and standard deviation of training and test accuracy as performance evaluation metrics.  Table \ref{tab:topnn} records the mean values and standard deviations of training and test accuracy of TopNN, ZeroNN, and NN across multiple datasets under varying-amplitude noise environments.  \\

\begin{table*}
    \centering
    \begin{NiceTabular}{r|P{1.2cm}P{1.2cm}|P{1.2cm}P{1.2cm}|P{1.2cm}P{1.2cm}|P{1.2cm}P{1.2cm}}
        \toprule
        Noise~~\, & \multicolumn{2}{c|}{None} & \multicolumn{2}{c|}{Weak} & \multicolumn{2}{c|}{Moderate} & \multicolumn{2}{c}{Strong} \\
        \midrule
        Accuracy & Training & Test & Training & Test & Training & Test & Training & Test \\
        \midrule[1pt]
        Dataset~\, & \multicolumn{8}{c}{ALLSSTAR} \\
        \midrule
        \RowStyle{\bfseries} TopNN & 91.2$\pm$0.2 & 91.2$\pm$0.2 & 90.7$\pm$0.5 & 90.5$\pm$0.5 & 89.6$\pm$0.3 & 89.5$\pm$0.4 & 88.6$\pm$0.4 & 88.1$\pm$0.5 \\
        ZeroNN                     & 90.5$\pm$0.2 & 90.2$\pm$0.3 & 89.5$\pm$0.5 & 89.3$\pm$0.4 & 88.3$\pm$0.5 & 88.4$\pm$0.5 & 87.0$\pm$0.5 & 86.7$\pm$0.6 \\
        NN                         & 90.4$\pm$0.3 & 90.2$\pm$0.4 & 89.6$\pm$0.4 & 89.4$\pm$0.4 & 88.4$\pm$0.4 & 88.3$\pm$0.5 & 87.0$\pm$0.4 & 86.7$\pm$0.6 \\
        \midrule[1pt]
        Dataset~\, & \multicolumn{8}{c}{LJSpeech} \\
        \midrule
        \RowStyle{\bfseries} TopNN & 94.1$\pm$0.2 & 94.1$\pm$0.2 & 91.6$\pm$0.6 & 91.5$\pm$0.4 & 90.5$\pm$0.7 & 90.5$\pm$0.8 & 88.4$\pm$0.7 & 88.1$\pm$0.6 \\
        ZeroNN                     & 93.4$\pm$0.2 & 93.3$\pm$0.3 & 89.5$\pm$2.6 & 89.3$\pm$2.4 & 88.8$\pm$2.0 & 88.8$\pm$2.0 & 86.1$\pm$2.9 & 85.3$\pm$3.0 \\
        NN                         & 93.3$\pm$0.3 & 93.3$\pm$0.3 & 89.6$\pm$2.6 & 89.4$\pm$2.6 & 88.6$\pm$1.8 & 88.7$\pm$1.9 & 85.0$\pm$3.7 & 84.3$\pm$3.6 \\
        \midrule[1pt]
        Dataset~\, & \multicolumn{8}{c}{TIMIT} \\
        \midrule
        \RowStyle{\bfseries} TopNN & 96.1$\pm$0.2 & 96.1$\pm$0.2 & 94.5$\pm$0.4 & 94.4$\pm$0.4 & 93.0$\pm$0.7 & 92.8$\pm$0.7 & 88.5$\pm$0.4 & 88.4$\pm$0.4 \\
        ZeroNN                     & 95.7$\pm$0.3 & 95.6$\pm$0.3 & 92.5$\pm$0.6 & 92.7$\pm$0.5 & 89.4$\pm$3.1 & 89.2$\pm$3.1 & 85.4$\pm$6.1 & 85.3$\pm$6.1 \\
        NN                         & 95.7$\pm$0.3 & 95.7$\pm$0.2 & 92.5$\pm$0.6 & 92.6$\pm$0.6 & 89.3$\pm$3.1 & 89.1$\pm$3.1 & 85.1$\pm$6.0 & 84.9$\pm$6.0 \\
        \bottomrule
    \end{NiceTabular}
    \caption{
        \textbf{Performance of state-of-the-art neural networks with topology enhancement.}  
        The table shows training and test accuracy rates of TopNN (in bold), ZeroNN (NN features concatenated with null topological feature, as a sanity check), and NN on original and noisy data across various datasets.  Levels of Gaussian white noise include none, weak (SNR\,=\,10dB), moderate (SNR\,=\,5dB), and strong (SNR\,=\,0dB).  All values are shown as mean\,$\pm$\,standard deviation, in percentage units \%.  The numerics are in supplement to the graphic demonstration in Fig.~\ref{fig:topnn}c--e and in partial comparison with the fifth and fourth rows from bottom of Table \ref{tab:topcap}.  These results demonstrate that TopNN achieves higher accuracy, steadier performance, and more robustness against noise.  
    }
    \label{tab:topnn}
\end{table*}

\noindent
Fig.~\ref{fig:topnn} shows that the training and test accuracy rates of TopNN consistently outperform those of ZeroNN and NN, with the latter two showing similar performance as expected.  When noise intensity increases, the performance gap between TopNN and the other two models widens, highlighting the former's robustness.  Additionally, TopNN exhibits lower accuracy variance across multiple experiments, indicating enhanced model stability.  

The results demonstrate that TopNN outperforms NN in classifying both clean and noise-injected speech data.  These findings collectively suggest that our proposed TopNN architecture achieves improved classification accuracy and robustness compared to the conventional NN framework.  The performance improvement can be attributed to the following synergistic mechanisms.  

The first mechanism concerns hybrid encoder.  Neural networks exhibit greater parametrisation flexibility and higher model complexity compared to fixed analytical paradigms, enabling task-specific feature extraction with enhanced generalisation capability.  However, their representational capacity for capturing intrinsic data structures remains constrained.  The topological approach complements this limitation by extracting multi-scale persistent homology features that are inherently difficult for neural networks to learn.  In this way, it enhances the representational capacity of the model.  

The second mechanism concerns hybrid decoder.  The proposed architecture employs fully connected layers as the decoder to dynamically learn the weights corresponding to the fused features.  This hybrid strategy exploits the strength of neural networks in learning hierarchical patterns while preserving the interpretability of topological descriptors.  

The third mechanism concerns enhanced robustness.  Neural networks, as data-driven models, are susceptible to performance degradation under limited or noisy training data.  In contrast, topological methods, such as persistent homology, focus on topological features (e.g., connected components, loops, voids) that persist across a range of scales rather than being sensitive to small local fluctuation.  This ensures that minor noise or perturbation in the data do not significantly alter the extracted topological features.  Therefore, the topological features extracted through topological methods demonstrate remarkable resistance to noise.  Our quantitative stability analysis as recorded in Table \ref{tab:topnn} confirms that the integrated framework with input of topological features significantly reduces variance in prediction outcomes and exhibits superior performance in classification tasks on noise-corrupted data.

\subsection*{Detection of vibration patterns}

The impetus behind TopCap lies in an observation of how PD can capture vibration patterns within time series.  To begin with, our aim is to determine which sorts of information can be extracted using topological methods.  As the name indicates, topological methods quantify features based on topology, which distinguishes spaces that cannot continuously deform to each other.  In the context of time series, we conduct a series of experiments to scrutinise the performance of topological methods, underscoring their limitation as well as their potential.  

Given a periodic time series, its TDE target is located on a closed curve (i.e., a loop) in a Euclidean space of sufficiently high dimension (see Fig.~\ref{fig:ph}a).  Despite the satisfactory point-cloud representation of a periodic time series, it remains rare in practical measurement and observation to capture a truly periodic series.  Often, researchers deal with time series that are not periodic yet exhibit certain patterns within some time segments.  

For instance, Fig.~\ref{fig:ph}c portrays the average temperature of the United States from the year 2012 to 2022, as documented in \cite{national_average_temp}.  Although the temperature does not strictly adhere to a periodic pattern, it does display a noticeable cyclic trend on an annual basis.  Typically, the temperature tends to rise from January to July and fall from August to December, with each year approximately comprising one cycle of the variation pattern.  A strength of topological methods is their ability to capture cycles.  A question then arises naturally: Can these methods also capture the cycle of temperature as well as subtle variation within and among these cycles?  

To be more precise, we observe that variation occurs in several ways.  For instance, the amplitude (or range) of the annual temperature variation may fluctuate slightly, with the maximum and minimum annual temperatures varying from year to year.  Additionally, the trend line for the annual average temperature also varies, such as the average temperature in 2012 surpassing that of 2013.  Despite each year's temperature pattern bearing resemblance to that depicted in the left panel in Fig.~\ref{fig:ph}c (representing a single cycle of temperature within a year), it may be more beneficial for prediction and response strategies to focus on the evolution of this pattern rather than its specific form.  In other words, attention should be directed towards how this cycle evolves over the years.  

This leads to several questions.  How can we consistently capture these subtle changes in the pattern's evolution, such as variation in frequency, amplitude, and trend line of cycles?  How can we describe the similarities and differences between time series that possess distinct evolutionary trajectories?  In applications, these are crucial inquiries that warrant further exploration.  

To address these questions, we propose three kinds of fundamental variations, which are utilised for depicting the evolutionary trace of a time series.  Let us consider a series of a periodic function $f(t_n)=f(t_n+T)$, where $T$ is a period.  

The first kind is variation of frequency.  Denote the frequency by $F=T^{-1}$.  Note that the series is not necessarily periodic in the mathematical sense.  Rather, it exhibits a recurring pattern after the period $T$.  For instance, the average temperature from Fig.~\ref{fig:ph}c is not a periodic series, but we consider its period to be one year since it follows a specific pattern, i.e., the one displayed in the left panel of Fig.~\ref{fig:ph}c.  This yearly pattern always lasts for a year as time progresses.  Hence, there is no frequency variation in this example.  This type of variation can be represented as $g_1(t_n)=f\big(F(t_n)\cdot t_n\big)$, where $F(t_n)$ is a series representing the changing frequency.  Such variation occurs, for example, when one switches their vocal tone or when one's heartbeats experience a transition from walking mode to running mode.  

The second kind is variation of amplitude.  The amplitudes of temperature in the years 2014 and 2015 are $42.73^\circ$F and $40.93^\circ$F, respectively.  So the variation of amplitude from 2014 to 2015 is $-1.80^\circ$F.  This can be represented by $g_2(t_n)=A(t_n)\cdot f(t_n)$, where $A(t_n)$ is a series of the changing amplitude.  This type of variation is observed when a particle vibrates with resistance or when there is a change in the volume of a sound.  

The third kind is variation of average line.  The average temperatures through the years 2012 and 2013 are $55.28^\circ$F and $52.43^\circ$F, respectively.  The variation of average line from 2012 to 2013 is $-2.85^\circ$F.  Let $g_3(t_n)=f(t_n)+L(t_n)$, where $L(t_n)$ is a series representing the variation of average line.  This type of variation is observed when a stock experiences a downturn over several days or when global warming causes a year-by-year increase in temperature.  

As a summary, Fig.~\ref{fig:ph}e provides a visual representation of the three fundamental variations.  It is important to note that these variations are not utilised to depict the pattern itself but rather to illustrate the variation within the pattern or how the time series oscillates over time.  This approach offers a dynamic perspective on the evolution of the time series, capturing changes in patterns that static analyses may overlook.  \\

\noindent
We first perform experiments on synthetic data.  

\begin{figure*}
    \centering
    \includegraphics[scale=.9]{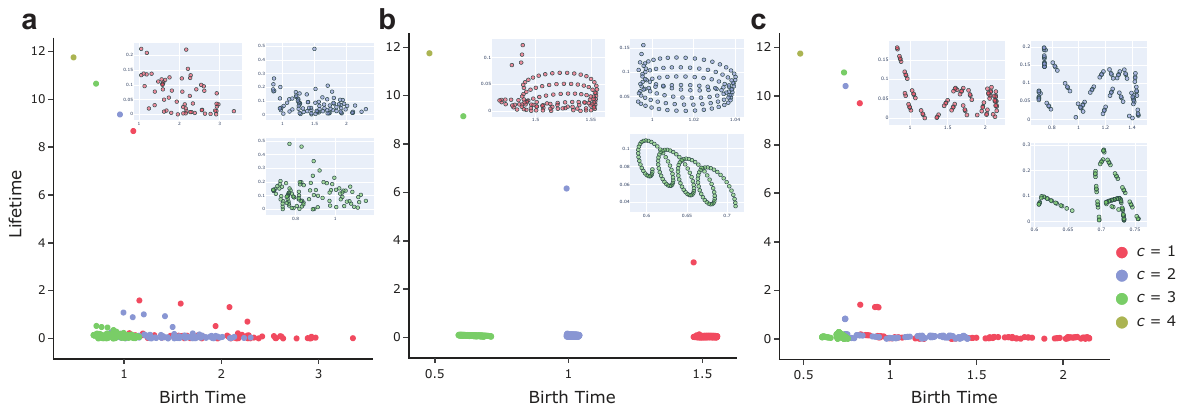}
    \caption{
        \textbf{1-Dimensional PH reveals three fundamental variations.}  
        In the colour legend, $c$ is the varying parameter in the synthetic time series.  
        \textbf{a} Detecting variation of frequency.  Upper-right panels zoom in to show the barcode distribution in the lower dense region, where the position and colour of each value of $c$ in the main legend corresponds to those of its panel.  Note that when $c=4$, there is a single point, and so the panel for this value is omitted.  
        \textbf{b} Detecting variation of amplitude.  
        \textbf{c} Detecting variation of average line.  
    }
    \label{fig:var-syn}
\end{figure*}

Let $t_n=0.01n$ with $0\leq t_n\leq 7\pi$, and for each $c\in\{1,2,3,4\}$ define 
\begin{equation*}
    \begin{split}
        f(t_n)   & =\cos(t_n) \\ 
        F(t_n)   & =\frac{c}{4}+\frac{1-\frac{c}{4}}{7\pi}\cdot t_n \\ 
        g_1(t_n) & =f\big(F(t_n)\cdot t_n\big) 
    \end{split}
\end{equation*}
Note that $F(t_n)=c/4$ when $t_n=0$ and $F(t_n)=1$ when $t_n=7\pi$.  In fact, $F(t_n)$ is a sequence of line segments connecting $(0,c/4)$ and $(7\pi,1)$.  Correspondingly, the frequency of $g_1(t_n)$ changes more slowly as $c$ increases.  In the extreme case when $c=4$, we have $F(t_n)=1$, and so 
\[
    g_1(t_n)=f\big(F(t_n)\cdot t_n\big)=f(t_n)=\cos(t_n) 
\]
which is a periodic function.  For each value of $c$, we apply TDE to the series $g_1(t_n)$ with a parameter choice of dimension 100, delay 3, skip 10 and compute the 1-dimensional PD of the embedded point cloud.  See Fig.~\ref{fig:var-syn}a for the results.  Replacing $F(t_n)$ by $A(t_n)$ and $L(t_n)$, we obtain the diagrams in Fig.~\ref{fig:var-syn}b and c, respectively.  

Using these three simulated time series corresponding to the three fundamental types of variation, we demonstrate that PD can distinguish these variations and detect how significant they are.  See Fig.~\ref{fig:var-syn}, where a smaller value of $c$ indicates a more rapid fundamental variation.  Here, regardless of which value $c$ takes, each individual diagram features a prominent single point at the top and a cluster of points with relatively short duration (or lifetime), except when $F(t_n)=1$ (i.e., $c=4$).  In this case, the series represents a cosine function, and thus the diagram consists of a single point.  

Normally, one tends to overlook the points in a PD that exhibit a short duration as they are sometimes inferred as noise.  However, in this example, the distribution of those points holds valuable information regarding the three fundamental variations.  As shown in Fig.~\ref{fig:var-syn}, each fundamental variation has its distinct pattern of distribution in the lower region of a diagram, which leads to refined inferences: If the points spiral along the vertical axis of lifetime, it is due to a variation of amplitude; if every two or four points stay close to form a bullet, it indicates a variation of average line; otherwise the points just seem to randomly spread over, which more likely results from a variation of frequency.  It is also straightforward to distinguish the values of $c$ for a specific fundamental variation, by their most significant point in the diagram.  Longer lifetime for the barcode of the solitary point indicates slower variation. The lower region of a diagram also gives evidence in this respect.  \\

\noindent
We next perform experiments on real-world data.  

In the previous simulated example, we demonstrated how PD could be utilised as a uniform means to distinguish three fundamental variations of the cosine series and their respective rates of change.  However, it is important to note that in general scenarios, identifying the fundamental variations in a time series using topological methods may encounter significant challenges.  Specifically, although topological methods are indeed capable of capturing this information, vectorising the topological descriptors for subsequent utilisation remains a complex task at this stage.  

Having recognised the potential of topological methods, we resort to an alternative algorithm for handling time series.  Despite the difficulty in vectorising PD to measure each fundamental variation, we develop a simplified algorithm to measure the vibration of time series as a whole.  This approach provides a comprehensive understanding of the overall behaviour of a time series, bypassing the need for complex vectorisation.  

We experiment with two records of the vowel [\textipa{A}] sourced from the HT1 corpus from the ALLSSTAR dataset of SpeechBox \cite{SpeechBox}.  Specifically, we demonstrate the fundamental variations by comparing the PDs of (a) a record of [\textipa{A}] relatively unstable with respect to the fundamental variations and (b) a second record of the same vowel that is relatively stable.  Here for TDE, the dimension is fixed to be 100, and the delay is computed the same way as in TopCap.  

\begin{figure*}
    \centering
    \includegraphics[scale=2]{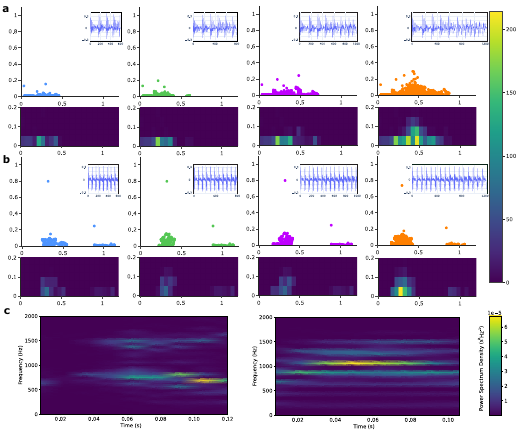}
    \caption{
        \textbf{Variation of 1-dimensional PDs due to the fundamental variations of time series.}  
        \textbf{a} PDs of drastic fundamental variations.  The small panel on top right of each diagram shows the original time series, with four segments extracted from the same record of [\textipa{A}], each starting from time 0 and ending at time 600, 800, 1000, 1200, respectively.  Below, each diagram shows the clustering density of points in the lower region of the PD.  
        \textbf{b} PDs of mild fundamental variations for four time-series segments extracted from the other record of [\textipa{A}], with the same ending and starting times as in (a).  It can directly be seen from the time series that the variation of amplitude in (a) is bigger than that in (b); for frequency, see (c); normally, we do not discuss the average line of phonetic data as it is assumed to be constant.  The lower density diagrams demonstrate that unstable time series are characterised by a higher density of points in the lower region of PD.  Moreover, stable series tend to attain high MP, consistent with Fig.~\ref{fig:var-syn}.  
        \textbf{c} Spectral frequency plots of the time series with drastic variation (left) and with mild variation (right).  The colour reflects amplitude.  
    }
    \label{fig:var-real}
\end{figure*}

To better illustrate the results, we crop each record into four overlapping intervals, each starting from time 0 and ending at 600, 800, 1000, 1200, respectively.  When adding a segment of 200 units into the original sample each time, the amplitude and frequency of the series altered more drastically in case (a).  A more rapid changing rate leads to more points distributed in the lower region of the diagram.  The results are presented in Fig.~\ref{fig:var-real}.  The plots in Fig.~\ref{fig:var-real}c show that the spectral frequency of (a) indeed varies faster than that of (b).  

We should also mention that the 1-dimensional PD here serves as a profile for the collective effect of the fundamental variations.  At this stage, unlike with synthetic data as in Fig.~\ref{fig:var-syn}, it is unclear how the points in the lower region change in response to a specific fundamental type of variation.

\section*{Discussion}

In this section, we present a comprehensive analysis of parameter selection strategies involved in the experiments above, and investigate challenges with their generalisation.  These strategies are geared towards both spectral and topological features of time series data.  
\\

\begin{figure*}
    \centering
    \includegraphics[scale=.95]{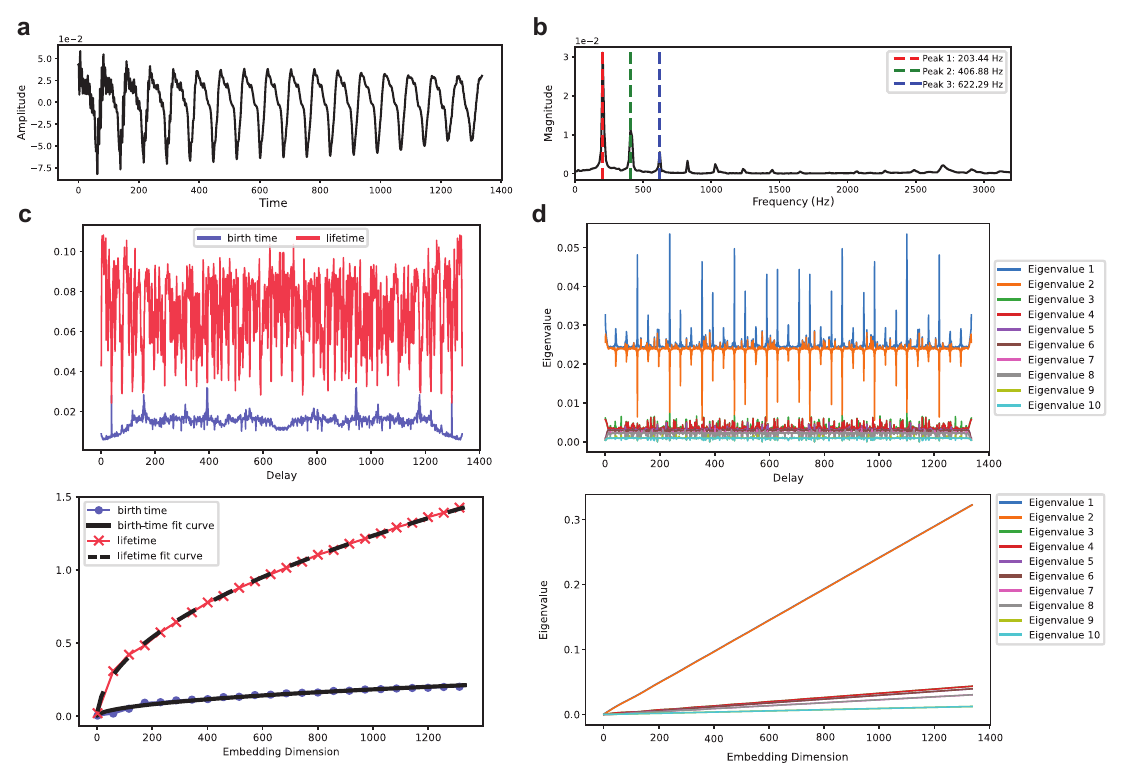}
    \caption{
        \textbf{Parameter selection and additional geometric features.}  
        \textbf{a} The original waveform diagram (.wav file) of the signal [\textipa{N}] (a voiced consonant) from the ALLSSTAR corpus.  
        \textbf{b} The power spectrum of the phone [\textipa{N}], with the first three prominent peaks annotated by red, green, and blue vertical dashed lines, corresponding to the first three formants (F1, F2, F3) in linguistic analysis.  The fundamental period of the speech signal can be derived from the frequency associated with Peak 1.  
        \textbf{c} Both the birth time and lifetime of maximal persistence via CTDE demonstrate extreme sensitivity with the delay parameter (upper, with fixed embedding dimension 10) and a smoothly proportional relationship following a square-root dependence on embedding dimension (lower, with fixed delay 10).  
        \textbf{d} The geometric distribution properties of time series via CTDE reveal some regular patterns in the first ten PCA eigenvalues.  Oscillation amplitude exhibits number-theoretic properties under varying delay parameter, and pairing with exponential decay (upper, with fixed embedding dimension 100), as well as linear proportional scaling with embedding dimension (lower, with fixed delay 10).  
    }
    \label{fig:disc}
\end{figure*}

\noindent
Central to our methods TopCap and TopNN is the TDE--PH pipeline for deriving the significant topological descriptor MP.  Given the Takens embedding theorem \cite{takens_detecting_1981,Broer-Takens}, the critical parameters of embedding dimension $d$ and time delay $\tau$ jointly govern the topological fidelity of reconstructed phase spaces.  We analyse from three aspects the interplay between $d$ and $\tau$ to elucidate their synergistic impact on optimising MP as follows.  

Firstly, we propose to solve the sample-size dilemma with large values of $d$ and $\tau$ by {\em cyclic} TDE.  Standard TDE imposes constraints on the minimal number of data points, requiring the number $N$ of data points to satisfy $N\geq(d-1)\cdot\tau$.  Moreover, PH analysis necessitates a significantly larger point cloud, demanding $N$ to be substantially greater than $(d-1)\tau$.  
    
However, in practical consonant recognition tasks, the finite length of speech data limits parameter exploration to a narrow range, as the maximal feasible $N$ is constrained by the inherent upper bound of audio duration.  To resolve this fundamental limitation and theoretically maximise the parameter search range for identifying optimal strategies, we propose a reconstruction method of {\em cyclic time-delay embedding} (CTDE).  By cyclically connecting the endpoints of the audio signal, CTDE enables multiples of $\tau$ and, equivalently, admissible values of $d$ and $\tau$ to ergodically traverse the entire interval $[1,N]$ of data points, utilising the full dataset without omission.  
    
It is worth noting that the number of embedded points remains $N$, independent of parameter choices, which yields a consistent and unbiased platform for systematic parameter optimisation.  Moreover, this approach does not compromise the discriminative properties for consonant classification.  For instance, given voiced consonants, which exhibit quasi-periodic structures, the cyclic reconstruction preserves their inherent periodicity.  For voiceless consonants, which resemble stochastic noise with uniformity and memorylessness, the endpoint connection maintains their statistical characteristics.  We give a more detailed discussion in Supplementary Sec.~\ref{subsec:ctde}, including 3D-projection visualisation of CTDE compared with TDE under varying parameters.  
    
Secondly, we find that MP correlates sublinearly to embedding dimension $d$.  As illustrated in the lower graph of Fig.~\ref{fig:disc}c, in our experiments with the voiced consonant [\textipa{N}], MP from CTDE exhibits a smooth nonlinear increase with respect to $d$, approximately following the relation MP $\propto d^{1/2}$.  This trend suggests that the growing MP scales sublinearly with embedding dimension.  

Combining our discussion on dependence of MP (from standard TDE) on $d$ in Supplementary Sec.~\ref{subsec:dim}, we see that the correlation between prominence of topological features and dimensionality stands in contrast to the common intuition from the curse of dimensionality as well as to the relatively low intrinsic dimensions of time series data.  Reasonably high embedding dimension improves overall performance of topology-enhanced ML.  
    
Thirdly, we observe sensitivity of MP to variation of time delay $\tau$.  In practice, MP exhibits extreme sensitivity to $\tau$, with its value oscillating violently under minor perturbations (see the upper graph of Fig.~\ref{fig:disc}c).  In contrast to the relationship between MP and $d$ discussed above, the one between MP and $\tau$ (with $d$ fixed) is highly discontinuous.  

In fact, Perea and Harer's assumptions break down for noisy or complex real-world time series, as the behaviour we observed contrasts sharply with that of idealised periodic signals (cf.~\cite{perea_sliding_2015}).  Under these experimental conditions, their conclusions predict that MP will attain maximal values at a discrete sequence $\tau=m\cdot T/d$, where $m$ are integers and $T$ is a fixed period of the time series.  As a result, the functional relation of MP with respect to $\tau$ must be a simple periodic function with each period $S$ containing exactly one maximum.  Moreover, the maximum value is invariant across successive periods, each being strictly $T/d$.  This limitation stems from a fundamental gap between its theoretical assumptions (e.g., strict periodicity) and the quasi-periodic nature of real-world signals such as human speech, where amplitude modulation and non-stationary dynamics dominate.  While the framework's parameter selection criteria as encoded in its closed-form equations may optimise alternative global geometric indices, such as geometric uniformity or spectral characteristics, these objectives are less well aligned with PH's focus on topological robustness, resulting in suboptimal MP performance.  
\\

\noindent
Now, let us discuss geometric distribution properties of time series embedded into high-dimensional space via CTDE from above.  

Principal component analysis (PCA) is a dimensionality reduction technique whose core objective is to project high-dimensional data into a low-dimensional space while preserving the primary structural information of the data.  Larger eigenvalues of the covariance matrix associated to the data indicate that the corresponding eigenvectors capture more significant variance in the data, meaning these directions are more informative and dominant in representing the underlying structure.  

Specifically, we investigate the case where the embedding dimension is fixed ($d=100$) while varying the delay parameter $\tau$ (through all possible values from 1 to $N-1$).  By sorting the eigenvalues in descending order and examining the top ten, we observe the following three patterns, as illustrated in Fig.~\ref{fig:disc}d.  

The first pattern is that oscillation amplitude exhibits number-theoretic properties.  Each eigenvalue oscillates with $\tau$, but its local average remains relatively stable over the global range.  This suggests that computing an average eigenvalue is meaningful, as only very few $\tau$ values deviate significantly from this average.  Spikes occur at {\em rational} points, i.e., where $\tau$ equals a rational multiple of the data length $N$.  Such values of $\tau$ lead to abrupt changes and sometimes even cause jumps to adjacent eigenvalues.  Although the example shown in the figure is not highly representative, for general audio signals, we observe that the amplitude of such mutations is negatively correlated with the denominator of the rational fraction.  The most significant changes occur at multiples such as $\frac{1}{2}$, $\frac{1}{3}$, $\frac{2}{3}$, $\frac{1}{4}$, $\frac{3}{4}$, etc.  

The second pattern is that the PCA eigenvalues exhibit pairing and exponential decay.  For each positive integer $k$, the average values of the $(2k-1)$\textsuperscript{st} and $(2k)$\textsuperscript{th} eigenvalues (sorted in descending order) are nearly identical, except for possible opposite jump directions at rational points.  Moreover, the magnitudes of the paired eigenvalues exhibit an exponential decay as $k$ increases.  

The third pattern concerns how random noise and high-frequency components affect the magnitude of PCA eigenvalues.  By introducing additive Gaussian white noise or substituting different audio files, we observe that higher randomness leads to mutations at more rational points, with larger amplitudes.  Similarly, a greater presence of high-frequency components in the Fourier spectrum results in more erratic behaviour.  

We next study the case where $\tau$ is fixed and $d$ varies.  The observed pattern is straightforward: Each eigenvalue grows linearly with $d$, but the growth rates differ, in a manner similar to the second pattern above.  

Finally, given the preceding discussion, we propose the closely related {\em formant spectral features} and {\em embedding configuration eigenvalues} as additional features for distinguishing voiced and voiceless consonants.  

In traditional linguistics and speech engineering, formant spectral features provide a relatively effective characterisation of phonemes, but their applicability has clear limitation.  According to \cite{vowel}, the first three formants (F1, F2, F3, represented by three differently coloured dashed lines in Fig.~\ref{fig:disc}b) can effectively explain the acoustic classification of vowels and voiced consonants.  However, they fail for voiceless consonants and are susceptible to coarticulation interference.  By using the frequencies and power intensities of Peak 1, Peak 2, and Peak 3 to form a 6-dimensional feature, we obtain classification accuracy rates of 93.5\% and 94.1\% for classifying voiced and voiceless consonants on the LJSpeech and TIMIT datasets, respectively.  

In our study of CTDE geometric configuration and PCA eigenvalues above, we discovered that the eigenvalues oscillate around stable mean values as the delay parameter $\tau$ varies, and together they serve as a robust invariant.  These eigenvalues are independent of $\tau$ and scale proportionally with the embedding dimension $d$, making them a potential feature for characterising intrinsic audio properties.  When applied to the same voiced/voiceless consonant classification task on the LJSpeech and TIMIT datasets, they yield accuracy rates of 88.1\% and 87.2\%, respectively.  As such, independent of PH, this approach provides a feature worthy of further investigation.  
\\

\noindent
In conclusion, the complexity of parameter selection in topological time series analysis lies in balancing theoretical models (e.g., the Perea--Harer framework) with non-periodic nature of real-world data.  While heuristics with fixed parameters offer pragmatic shortcuts, future work must focus on adaptive, signal-tailored frameworks.  Integrating dimensionality reduction, noise-robust persistence representation, and hybrid spectral--topological methods could unlock more reliable and generalisable solutions.

\section*{Data Availability}

The data that support the findings of this study are openly available in SpeechBox, ALLSSTAR Corpora \cite{SpeechBox}, as well as LJSpeech \cite{LJSpeech}, TIMIT \cite{TIMIT}, and LibriSpeech \cite{LibriSpeech}.

\section*{Code Availability}

The source code for TopCap and related models can be accessed at \cite{TopML}.

\section*{References}

\printbibliography[notkeyword=web,title={\hskip5cm}]
\newrefcontext[labelprefix=]

\section*{Acknowledgements}

The authors would like to thank Meng Yu of Tencent AI Lab for his mentorship on audio and speech signal processing, on neural networks and deep learning, as well as for his comments and suggestions on an earlier draft of this manuscript.  The authors would also like to thank Andrew Blumberg, Gunnar Carlsson, Fangyi Chen, Haibao Duan, Houhong Fan, Fuquan Fang, Yulia Gel, Yunan He, Wenfei Jin, Tyler Lawson, Fengchun Lei, Jingyan Li, Yanlin Li, Hongwei Lin, Andy Luchuan Liu, Tao Luo, Zhi Lü, Jianxin Pan, Yuhe Qin, Qi Sun, Yujiao Jennifer Sun, Guoliang Tian, Jie Wu, Rongling Wu, Kelin Xia, Tony Xiaochen Xiao, Jiang Yang, Jin Zhang, and Zhen Zhang for discussions and encouragement.

\section*{Funding Statement}

This work was partly supported by the National Natural Science Foundation of China grant 12371069, Guangdong Basic and Applied Basic Research Foundation grant 2023A1515030289, and the Guangdong Provincial Key Laboratory of Interdisciplinary Research and Application for Data Science (2022B1212010006, UICR0600008-7).

\section*{Author Contributions Statement}

Y.Z.~planned the project.  P.F.~and S.Y.~constructed the theoretical framework.  P.F., Q.Q., and H.Z.~designed the sample, built the algorithms, and analysed the data.  S.Y., Z.Y., and Z.D.~assisted with the algorithms and data analysis.  Z.D.~assisted with the code repositories.  P.F., Q.Q., H.Z., S.Y., Z.Y., Z.D., and Y.Z.~wrote the paper and contributed to the discussion.

\section*{Competing Interests Statement}

The authors declare no competing interests.  

\clearpage

\section*{Supplementary Information}

\vskip.5cm

\begin{center}
    \large\textsf{Topology-enhanced machine learning for speech signal processing} 
\end{center}

\renewcommand{\figurename}{Supplementary Figure}
\renewcommand{\tablename}{Supplementary Table}

\setcounter{figure}{0}
\setcounter{table}{0}

\section{Phonetic data, auditory perception, and learning topologically}
\label{sec:phone}

In this section, we first review the basics of phonetic data, our main objects of study, and explain our scientific approach towards a distribution space for them based on their topological features (rather than biomechanical production).  We then review the mechanism of human auditory perception, especially the structure of a cochlea as a biological Fourier analysis machine.  This underpins existing audio and speech signal processing technology.  In contrast, our topological approach to phonetic data extends beyond biomimetic engineering to more comprehensive, robust feature extraction and learning, as demonstrated with TopNN from Results in the main text.

\subsection{Phonetic data and their distribution}

As a research field of linguistics, phonetics studies the production as well as the classification of human speech sounds from the world's languages.  In phonetics, a {\em phoneme} is the smallest basic unit of human speech sounds.  It is a short speech segment possessing distinct physical or perceptual properties.  In the main text and Supplementary Information, to differentiate the theoretical and real-world objects, we reserve the word {\em phone} for a phoneme segmented from a recording of human speech.  

Phonemes are generally classified into two principal categories: vowels and consonants.  A {\em vowel} is defined as a speech sound pronounced by an open vocal tract with no significant build-up of air pressure at any point above the glottis, and at least making some airflow escape through the mouth.  In contrast, a {\em consonant} is a speech sound that is articulated with a complete or partial closure of the vocal tract and usually forces air through a narrow channel in one's mouth or nose.  

Unlike vowels which must be pronounced by vibrated vocal cords, consonants can be further categorised into two classes according to whether the vocal cords vibrate or not during articulation.  If the vocal cords vibrate, the consonant is known as a {\em voiced} consonant.  Otherwise, the consonant is {\em voiceless} (or {\em unvoiced}).  Since vocal cord vibration can produce a stable periodic signal of air pressure, voiced consonants tend to have more periodic components than voiceless consonants, which can in turn be detected by PH as topological characteristics from phonetic time series data.  

Indeed, one of the more heuristic motivations for our research project is to re-examine (and even revise) the linguistic classifications of phonemes through the mathematical lens of topological patterns and shape of speech data, analogous to Carlsson and his collaborators' seminal work \cite{Scarlsson_local_2008} on the distribution of image data (cf.~the distribution chart of vowels created by linguists in \cite{SIPA}).

\subsection{Spectral signal processing and beyond}

The transmission of sound to the human auditory system is a marvel of biological engineering, where acoustic waves are progressively transformed into neural signals.  This process begins with the external ear channelling sound waves to the tympanic membrane, which subsequently induces vibration in the ossicles of the middle ear (the malleus, incus, and stapes, constituting the smallest bones in the human body).  These minute oscillations are then conveyed to one of the most critical structures in auditory perception: the cochlea.  

The cochlea, in essence, functions as a biological Fourier analysis machine (see \cite{SEAR} for an illustration).  This spiral-shaped, fluid-filled organ amplifies the incoming sound waves and performs a spectral decomposition of complex acoustic signals.  The cochlea's architecture is characterised by a gradual variation in the radius of its spiral and the corresponding mechanical properties of the basilar membrane that runs along its length.  The basal end of the cochlea, with its rigid basilar membrane and narrow duct, is optimally tuned to high-frequency vibrations.  In contrast, the apical region, featuring a more flexible membrane and wider duct, is more responsive to signals of lower frequency.  

This structural gradient creates a tonotopic organisation within the cochlea, analogous to the varying tensions of musical strings producing different pitches.  The basilar membrane's varying mechanical properties result in different regions having distinct resonant frequencies, each maximally sensitive to a specific range of sound frequencies.  Atop this membrane reside the hair cells, specialised mechanoreceptors that transduce mechanical vibrations into electrical signals and enable auditory perception.  

The cochlea's spiral configuration, in conjunction with the basilar membrane's properties, constitutes a natural, passive mechanical Fourier analyser.  This biological mechanism effectively distributes frequency components of sound waves along the length of the cochlea.  Consequently, the neural signals generated by hair cells at different locations along the basilar membrane correspond to distinct frequency bands of the original acoustic input.  

It is noteworthy that contemporary industrial approaches to speech signal processing, such as STFT and MFCC as compared in Results of the main text, employ analytical methods that parallel the cochlea's function.  These techniques decompose signals into linear combinations of basis functions, mirroring the cochlea's spectral analysis.  This convergence of biological design and signal processing methodology can be viewed as a triumph of biomimetic engineering.  

Intriguingly, our experimental findings have demonstrated that topological principles can also be leveraged to extract certain acoustic information.  This approach lacks a direct physiological counterpart in current auditory research and established theoretical frameworks.  The potential for topological methods in auditory signal processing opens up an exciting frontier for exploration, bridging the gap between abstract mathematics and biological sensory systems.  Future investigation in this domain may yield insights that could transform our understanding of auditory perception and inspire innovative signal processing techniques (cf.~\cite{SRingach,Stopo_sig,Sgestalt}).

\section{Persistent homology}
\label{sec:ph}

Topology is a subject that studies the properties of geometric objects that remain unchanged under continuous transformations or smooth perturbations.  It focuses on the intrinsic features of a space regardless of its rigid shape or size.  Algebraic topology provides a quantitative description of these topological properties.

\subsection{Simplicial complexes}

The structure of a {\em simplicial complex} (and its numerous variants and analogues) provides a powerful tool in algebraic topology which enables us to represent a topological space using discrete data.  Unlike the original space, which can be challenging to compute and analyse, a simplicial complex provides a combinatorial description that is much more amenable to computation.  We can use algebraic techniques to study the properties of a simplicial complex, such as its homology and cohomology groups, which encode and reveal information about the topology of the represented space.  

Formally, a simplicial complex with {\em vertices} in a set $V$ is a collection $K$ of nonempty finite subsets $\sigma\subset V$ such that any nonempty subset $\tau$ of $\sigma$ is also contained in $K$ (called a {\em face} of $\sigma$) and that $\sigma,\sigma'\in K$ implies their intersection is either empty or a face of both.  A set $\sigma\in K$ with $(i+1)$ elements is called an {\em $i$-simplex} of the simplicial complex $K$.  

\begin{figure}[h]
    \centering
    \includegraphics[scale=.9]{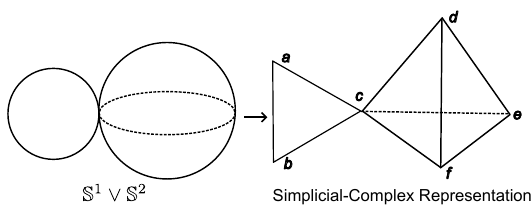}
    \caption{\textbf{From a topological space to its triangulation.}}
    \label{fig:simplex}
\end{figure}

For instance, Supplementary Fig.~\ref{fig:simplex} illustrates $\mathbb{S}^1\vee\mathbb{S}^2$, a circle kissing a sphere at a single point.  This topological space can be represented by a simplicial complex $K$ with six vertices $a$, $b$, $c$, $d$, $e$, and $f$, namely, 
\[
    \begin{split}
        K=\big\{ & \{a\},\{b\},\{c\},\{d\},\{e\},\{f\}, \\ 
                 & \{a,b\},\{a,c\},\{b,c\},\{c,d\},\{c,f\},\{d,f\},\{c,e\}, \\ 
                 & \{d,e\},\{f,e\}, \\
                 & \{c,d,f\},\{c,e,f\},\{c,d,e\},\{d,e,f\}\big\} 
    \end{split}
\]
It is a combinatorial avatar for $\mathbb{S}^1\vee\mathbb{S}^2$ via a triangulation operation on the latter, which is amenable to algebraic computation as follows.

\subsection{Homology groups and Betti numbers}

Let $p$ be a prime number and $\mathbb{F}_p$ be the field with $p$ elements.  Given a simplicial complex $K$, define $C_i(K;\mathbb{F}_p)$ to be the $\mathbb{F}_p$-vector space with basis the set of $i$-simplices in $K$.  To keep track of the order of vertices within a simplex, we use the alternative notation with square brackets in the following.  If $\sigma=[v_0,v_1,\ldots,v_i]$ is an $i$-simplex, define the {\em boundary} of $\sigma$, denoted by $\partial\sigma$, to be the alternating sum of the $(i-1)$-dimensional faces of $\sigma$ given by 
\[
    \partial\sigma\coloneqq\sum_{k=0}^i(-1)^k[v_0,\ldots,\widehat{v_k},\ldots,v_i] 
\]
where $[v_0,\ldots,\widehat{v_k},\ldots,v_i]$ is the $k$\textsuperscript{th} $(i-1)$-dimensional face of $\sigma$ missing the vertex $v_k$.  We can extend $\partial$ over $C_i(K;\mathbb{F}_p)$ as an $\mathbb{F}_p$-linear operator so that $\partial\colon C_i(K;\mathbb{F}_p)\to C_{i-1}(K;\mathbb{F}_p)$.  The composition of boundary operators satisfies $\partial\circ\partial=0$.  The elements in $C_i(K;\mathbb{F}_p)$ with boundary 0 are called {\em $i$-cycles}.  They form a subspace of $C_i(K;\mathbb{F}_p)$, denoted by $Z_i(K;\mathbb{F}_p)$.  The elements in $C_i(K;\mathbb{F}_p)$ that are the images under $\partial$ of elements in $C_{i+1}(K;\mathbb{F}_p)$ are called {\em $i$-boundaries}.  They form a subspace too, denoted by $B_i(K;\mathbb{F}_p)$.  It follows from $\partial\circ\partial=0$ that 
\[
    B_i(K;\mathbb{F}_p)\subset Z_i(K;\mathbb{F}_p) 
\]
Then define the quotient space 
\[
    H_i(K;\mathbb{F}_p)\coloneqq Z_i(K;\mathbb{F}_p)/B_i(K;\mathbb{F}_p) 
\]
to be the {\em $i$\textsuperscript{th} homology group of $K$ with $\mathbb{F}_p$-coefficients}.  We call ${\dim}\big(H_i(K;\mathbb{F}_p)\big)$ the {\em $i$\textsuperscript{th} Betti number}, denoted by $\beta_i(K)$, which counts the number of $i$-dimensional loops (bounding an $(i-1)$-dimensional void) in the corresponding topological space.  As such, these homology groups are also called the homology groups of the space (it can be shown that they are independent of the particular ways in which the space is triangulated).  For example, the Betti numbers of $\mathbb{S}^1\vee\mathbb{S}^2$ above are $\beta_1=1$, $\beta_2=1$, and $\beta_i=0$ when $i\geq3$.

\subsection{Persistent homology and its representations}

The usefulness of these invariants, besides their computability (essentially Gaussian elimination in linear algebra), lies in their tractability along deformations.  Given two simplicial complexes $K$ and $L$, a simplicial map $f\colon K\to L$ (that preserves the simplicial structure) induces an $\mathbb F_p$-linear map $H_i(f;\mathbb{F}_p)\colon H_i(K;\mathbb{F}_p)\to H_i(L;\mathbb{F}_p)$.  Thus, if two spaces are topologically equivalent (in fact, being homotopy equivalent suffices), their homology groups must be isomorphic and the Betti numbers match up.  

Let $(X,d)$ be a finite point cloud with metric $d$.  Define a family of simplicial complexes, called {\em Rips complexes}, by 
\[
    R_\epsilon(X)\coloneqq\{\sigma\subset X\,|\,d(x,x')\leq\epsilon~\text{for all}~x,x'\in\sigma\} 
\]
where $\epsilon\geq0$ is a parameter.  The family 
\[
    \mathcal{R}(X)\coloneqq\{R_{\epsilon}(X)\}_{\epsilon\geq0} 
\]
is known as the Rips filtration of $X$.  Clearly, if $\epsilon_1\leq\epsilon_2$, then $R_{\epsilon_1}(X)\hookrightarrow R_{\epsilon_2}(X)$.  Thus, for each $i$ we obtain a sequence 
\[
    \begin{split}
        H_i\big(R_{\epsilon_0}(X);\mathbb{F}_p\big)\to H_i\big(R_{\epsilon_1}(X);\mathbb{F}_p\big) & \to\cdots \\
        & \to H_i\big(R_{\epsilon_m}(X);\mathbb{F}_p\big) 
    \end{split}
\]
where $0=\epsilon_0<\epsilon_1<\cdots<\epsilon_m<\infty$.  As $\epsilon$ varies, the topological features of the simplicial complexes $R_{\epsilon}(X)$ vary, resulting in the emergence and disappearance of loops (see Fig.~\ref{fig:ph}d).  

Given the values of $\epsilon$, record the instances of emergence and disappearance of loops, which correspond to cycle classes in the homology groups along the above sequence.  Thus each class has a descriptor $(b,d)\in \mathbb{R}^2$, where $b$ represents its {\em birth time}, $d$ represents its {\em death time}, and $d-b$ represents its {\em lifetime}.  In this way, we obtain a multiset 
\[
    \{(b_j,d_j)\}_{j\in J}\eqqcolon{\rm dgm}_i\big(\mathcal{R}(X)\big) 
\]
which encodes the persistence of topological features of $X$.  This multiset can be represented as a multiset of points in the 2-dimensional coordinate system, called a {\em persistence diagram} for the $i$\textsuperscript{th} PH, or as an array of interval segments, called a {\em persistence barcode} (see Fig.~\ref{fig:ph}f).  In particular, by {\em maximal persistence} we refer to the maximal lifetime among all the points in a persistence diagram.  For convenience, sometimes we adopt a variant of the persistence diagram whose vertical axis represents lifetime instead of death time, such as in Fig.~\ref{fig:var-syn}a.

\section{Time-delay embedding}
\label{sec:tde}

Time-delay embedding (TDE) is also known as Takens's embedding, sliding window embedding, delay embedding, and delay coordinate embedding.  For simplicity, we focus on 1-dimensional time series.

\subsection{Continuous time series}
\label{subsec:cts}

TDE of a real-valued function $f\colon\mathbb{R}\to \mathbb{R}$, with parameters positive integer $d$ and positive real number $\tau$, is defined to be the vector-valued function 
\[
    \begin{split}
        \mathcal{E}_{d,\tau}f\colon\mathbb{R} & \to\mathbb{R}^d \\
        t & \mapsto\Big(f(t),f(t+\tau),\ldots,f\big(t+(d-1)\tau\big)\Big) 
    \end{split}
\]
Here, $d$ is the {\em dimension} of the target space for the embedding, $\tau$ is the {\em delay}, and their product $\tau\cdot d$ is called the {\em window size}.  

Let $M$ be a compact smooth manifold of dimension $m$, $\{\phi_t\colon M \to M\}_{t>0}$ be a smooth dynamical system, and $G\colon M\to\mathbb{R}$ be a smooth (observation) function.  According to Takens's embedding theorem \cite{Stakens_detecting_1981}, for generic pairs $(\phi_t,G)$ and a fixed $\tau>0$, the delay-coordinate map $\Psi\colon M\to\mathbb{R}^d$ defined by 
\[
    \Psi(x)=\Big(G(x),G\big(\phi_\tau(x)\big),G\big(\phi_{2\tau}(x)\big),\ldots,G\big(\phi_{(d-1)\tau}(x)\big)\Big) 
\]
is an embedding, provided that $d\geq2m+1$.  

Therefore, with the assumption that our time series data is generated by an unknown dynamical system evolving on a smooth manifold and an unknown observation function, i.e., $f(t)=G\big(\phi_t(x_0)\big)$, the image $\mathcal{E}_{d,\tau}f$ of the TDE reconstructs the topological shape of the trajectory of the base point $x_0$ in the manifold $M$ up to homeomorphism, provided the condition $d\geq2m+1$.  In particular, when the trajectory converges to an attractor, the reconstruction quality improves significantly.  This is because attractors are invariant sets, in the sense that once a trajectory enters an attractor, it remains within it, and nearby trajectories asymptotically approach it.  Moreover, attractors are minimal, that is, they cannot be decomposed into smaller invariant subsets.  Consequently, the reconstructed point cloud becomes denser near the attractor, giving a more faithful representation of the underlying dynamics.  

In \cite[Sec.~5]{Sperea_sliding_2015}, Perea and Harer established that the $n$-truncated Fourier series expansion 
\[
    S_nf(t)=\sum_{k=0}^na_k\cos(kt)+b_k\sin(kt) 
\]
of a periodic time series $f$ can be reconstructed into a circle when $d\geq2n$, i.e., 
\[
    \mathcal{E}_{d,\tau}S_nf(\mathbb{R})\cong\mathbb{S}^1 
\]
Moreover, let $L$ be a constant such that 
\[
    f\left(t+\frac{2\pi}{L}\right)=f(t) 
\]
Then the 1-dimensional MP of the resulting point cloud is the largest when the window size $\tau\cdot d$ is integrally proportional to $2\pi/L$, i.e., 
\[
    \tau\cdot d=m\cdot\frac{2\pi}{L} 
\]
for a positive integer $m$.  Intuitively, an increase in the embedding dimension results in a better approximation when truncating the Fourier series, and the MP of the point cloud becomes the most significant when the window size equals a period.  

This methodology also proves particularly advantageous in scenarios where the system under investigation exhibits nonlinear dynamics, precluding straightforward analysis of the time series data.  Via a suitable embedding, the inherent geometric configuration of the system emerges, enabling deeper comprehension and refined analysis.

\subsection{Discrete time series}

In practice, we work with discrete-time signals obtained by sampling a continuous function over a fixed time interval, which we often normalise to be of length $\Delta t=1$.  Suppose that the number of samples equals $N$, and write $[N]$ for the set $\{0,1,\ldots,N-1\}$.  Given a discrete signal $f\colon[N]\to\mathbb{R}$, we can apply TDE directly to the sequence, by taking the delay parameter $\tau$ to be a positive integer.  

It is also common to introduce an additional parameter of {\em skip} $s$, with $s$ a positive integer, that controls the step between successive starting points of windows.  In effect, with this skip we down-sample the signals by taking every $s$ windows (of size $\tau\cdot d$), reducing the amount of data and computational load.  This procedure can also act as a simple low-pass filter that attenuates high-frequency noise.  

Given skip $s$, the set $T_s$ of {\em admissible window starting points} consists of those sample times $t$ such that a window of $d$ points beginning at $t$ and evenly spaced by $\tau$ stays within the signal range $[N]$, and such that these times $t$ are evenly spaced by $s$.  Typically, a sampling begins with $t = 0$, so that 
\[
    T_s=\{t\in[N]\,|\,\text{$t+(d-1)\tau\leq N-1$ and $s$ divides $t$}\} 
\]
The corresponding TDE is then 
\[
    \begin{split}
        \mathcal{E}_{d,\tau,s}f\colon T_s & \to\mathbb{R}^d \\
        t & \mapsto\Big(f(t),f(t+\tau),\ldots,f\big(t+(d-1)\tau\big)\Big) 
    \end{split}
\]
It converts the signal $f$ into a point cloud in $\mathbb{R}^d$, with one $d$-dimensional vector for each window starting point $t\in T_s$.  

For example, let us consider a signal sequence $f(0),f(1),\ldots,f(9)$ with $N=10$, and choose embedding parameters $d=3$, $\tau=2$, and $s=3$ so that $T_s=\{0,3\}$.  The embedding then yields two points in $\mathbb{R}^3$, namely, 
\[
    \begin{split}
        \mathcal{E}_{3,2,3}f(0) & =\big(f(0),f(2),f(4)\big)~\text{and} \\
        \mathcal{E}_{3,2,3}f(3) & =\big(f(3),f(5),f(7)\big) 
    \end{split}
\]

\subsection{Cyclic time-delay embedding}
\label{subsec:ctde}

While standard TDE offers a fundamental approach for state space reconstruction, its parameter selection faces critical constraints requiring the sample size (i.e., number of sampled points) $N\geq(d-1)\tau$.  Given a discrete signal $f\colon[N]\to\mathbb{R}$, conventional TDE with parameters $d$ and $\tau$ generates at most $N-(d-1)\tau$ embedded points (with skip $s=1$), which creates a sample-size dilemma that severely restricts practical applications with small series length $N$ or large values of $d$ and $\tau$.  

To overcome this limitation, we propose a reconstruction method of {\em cyclic time-delay embedding} (CTDE).  By implementing cyclic boundary conditions through modular arithmetic, CTDE preserves the complete dataset without truncation.  Our method enables nearly unrestricted parameter selection, allowing $d$ and $\tau$ to independently vary over the full parameter space from 1 to $N$.  Crucially, CTDE maintains a sample size of $N$ embedded points regardless of parameter choices (if the skip $s=1$), which yields a consistent and unbiased platform for systematic parameter optimisation.  Formally, the CTDE mapping is defined as follows: 
\[
    \begin{split}
        \mathcal{CE}_{d,\tau,s}f\colon T_s & \to\mathbb{R}^d \\
        t & \mapsto\Big(f(t\bmod N),\,f(t+\tau\bmod N),\,\ldots, \\
        & \hskip.75cm f\big(t+(d-1)\tau\bmod N\big)\Big) 
    \end{split}
\]
where $r=a\mod N$ denotes the standard modulo operation that returns the unique integer $r\in[0,N)$ satisfying $a=kN+r$ for some integer $k$.  

\begin{figure*}
    \centering
    \includegraphics[scale=.76]{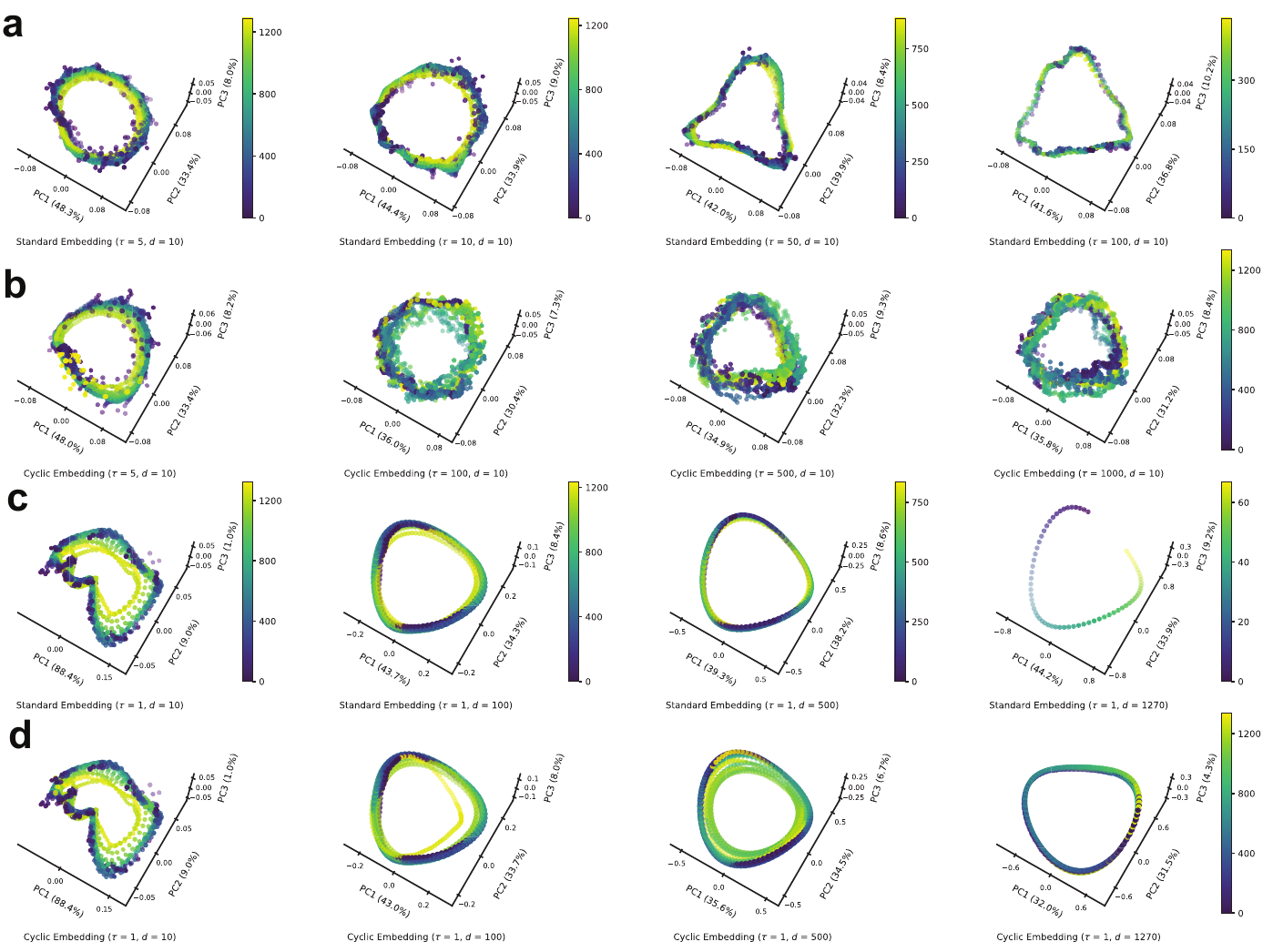}
    \caption{
        \textbf{Visualisation of embedded point clouds via standard TDE and cyclic TDE.}  
        The plots show PCA projections of the embedded point clouds in 3D with various embedding dimensions and time delays.  In each plot, the percentages along an axis represent the explained variance ratio of a PCA eigenvalue.  The colour legends show locations of sampled data points in the time series prior to embedding.  Standard TDE yields sparse and scattered points when both embedding dimension $d$ and time delay $\tau$ are large.  The number of sampled points that get embedded is constrained by these parameters, hence the varying colour legends in (a) and (c).  
        \textbf{a} TDE with fixed $d=10$.  Time delay $\tau$ varies over $\{5,10,50,100\}$, but cannot increase significantly further.  
        \textbf{b} CTDE with fixed $d=10$.  Time delay $\tau$ varies over a bigger range including $\{5,100,500,1000\}$, which demonstrates greater flexibility.  
        \textbf{c} TDE with fixed $\tau=1$.  Embedding dimension $d$ varies over $\{10,100,500,1270\}$.  Notably, when $d$ reaches 1270, the point cloud gaps, without forming a closed cycle (i.e., head-to-tail connection).  Consequently, no significant 1-dimensional MP can be captured in the phonetic time series.  In fact, when the embedding dimension further increases to 1290, an empty 1-dimensional barcode is obtained due to the lack of points necessary to form even a single cycle.  
        \textbf{d} CTDE with fixed $\tau=1$.  Embedding dimension $d$ again varies over $\{10,100,500,1270\}$, which showcases the method's improved entirety and faithfulness in representing topological characteristics.  
    }
    \label{fig:tde}
\end{figure*}

Supplementary Fig.~\ref{fig:tde} shows embedded point clouds generated by both standard and cyclic TDE methods, displaying their 3D PCA projections across various embedding dimensions and time delays.  The sample of a voiced consonant is the same as the one in Fig.~\ref{fig:disc}a.  

\section{Methods with TopCap}
\label{sec:method}

In this section, we provide details of some of the methods applied when carrying out the experiments in the main text, particularly TopCap (cf.~Fig.~\ref{fig:topcap}e).

\subsection{Deriving phonetic data from natural speech}
\label{subsec:m_pre}

We use natural-speech files sourced from the ALLSSTAR Corpus of SpeechBox \cite{SSpeechBox}, task HT1 language English L1, retrieved on 28th January 2023.  SpeechBox is a web-based system providing access to an extensive collection of digital speech corpora developed by the Speech Communication Research Group in the Department of Linguistics at Northwestern University.  

The HT1--English section contains a total of 25 individual files, comprising 14 files from women and 11 files from men.  The age range of these speakers spans from 18 to 26 years, with an average of 19.92.  Each file is presented in the WAV format and is accompanied by its corresponding aligned file in Textgrid format, which features three tiers of sentences, words, and phones.  Collectively, these 25 speech files amount to a total duration of 41.21 minutes.  Each speech file contains an individual reading the same sentences consecutively for a duration ranging from 80 to 120 seconds, contingent upon their pace.  

The original .wav file has a sampling frequency of 22050 Hz and comprises only one channel.  Since MFA \cite{Smcauliffe17_interspeech} is trained in a sampling frequency of 16 kHz, we opt to adjust the sampling frequency of the .wav files accordingly.  We then extract the words tier from Textgrid and align words into phones using English\_MFA dictionary and acoustic model (MFA version 2.0.6).  Thus we obtain corresponding phonetic data from these speech files.  

Subsequently, we use voiced and voiceless consonants in those segments as our dataset.  Voiced consonants are consonants for which vocal cords vibrate in the throat during articulation, while voiceless consonants are pronounced otherwise (see Supplementary Sec.~\ref{sec:phone} for more details).  Specifically, using Praat \cite{SPraat}, we extract voiced consonants [\textipa{N}], [m], [n], [j], [l], [v], and [\textipa{Z}].  For voiceless consonants, we select [f], [k], [\textipa{8}], [t], [s], and [\textipa{tS}].  Our selection is limited to these voiced and voiceless consonants, as we aim to balance the ratio of voiced and voiceless consonant records in these speech files.  Additionally, some consonants, such as [d] and [h], appear difficult to classify by our methods.

\subsection{Selecting parameters for time-delay embedding}
\label{subsec:m_tde}

Before extracting topological features from a time series, we first endow this 1-dimensional time series with a (Euclidean) topological structure through TDE (see Supplementary Sec.~\ref{sec:tde}).  It is noteworthy that this technique also applies to multi-dimensional time series.  The ambient space throughout this article is always a Euclidean space.  By establishing the topological structure there, or more precisely, the distance matrices, we subsequently compute PH.  

For TDE, we select suitable parameters to capture the theoretically optimal MP of a given time series.  Here, the embedding dimension is fixed to be 100.  Our principle for determining an appropriate dimension is that we want to choose the embedding dimension to be large for a time series of limited length.  As discussed in Supplementary Sec.~\ref{sec:tde} and Discussion in the main text, a higher dimension results in a more accurate approximation by sampled data points.  This approach also aims at enhancing computational efficiency and the occurrence of more prominent MP.  

Nonetheless, it is important to exercise caution when selecting the dimension, as excessively large dimensions may lead to empty point clouds and other uncontrollable factors.  For instance, with a time series consisting of approximately 1200 points, setting the dimension to 100, delay to 5, and skip to 1 results in around 700 points in the corresponding point cloud in Euclidean space.  However, increasing the dimension to 200 with the same remaining parameters yields only 200 points, which may be less sufficient to adequately represent the original data structure.  Thus, the dimension is chosen to be as large as possible while maintaining enough data points in the point cloud.  

With a proper dimension, we then compute the delay parameter for the embedding.  According to Perea and Harer \cite{Sperea_sliding_2015}, in the case of a periodic function, the optimal delays $\tau$ can be expressed as 
\begin{equation}
\label{tau}
    \tau=m\cdot\frac{T}{d} 
\end{equation}
where $T$ denotes the (minimal) period, $d$ represents the dimension of the embedding, and $m$ is a positive integer.  

Under these conditions, we obtain the theoretically optimal MP.  However, the time series under consideration in our experiments are far from periodic.  As a result, we use the first peak of the ACL function to represent the period $T$.  The common choice of $\tau$ is to let the window size $\tau\cdot d$ equal the (minimal) period.  However, in the case of a discrete time series, one often obtains $\tau=0$ or $\tau=1$ in this way, since the embedding dimension $d$ is too large in comparison.  Therefore, one strategy is to increase $m$ to get a relatively reasonable $\tau$.  Here we set $m=6$.  Then the resulting $\tau$ from Supplementary Equation \eqref{tau} is rounded to the nearest integer (if it equals 0, take 1 instead).  The performance of MP with delay obtained in this way is presented in Supplementary Sec.~\ref{subsec:mp} below.  

It is not uncommon that the window size exceeds the number of points in a time series, resulting in an empty embedding.  Let $|S|$ denote this number of (sampled) points, i.e., the data-point capacity of the time series.  In the extreme case where $|S|<500$, we apply our variant of cyclic TDE (see Supplementary Sec.~\ref{subsec:ctde} and Discussion in the main text), with $d$ and $\tau$ chosen as above.  Otherwise, if $500\leq|S|\leq\tau\cdot d$, we adopt the delay $\tau'=|S|/d$ instead, rounded downwards.  These measures enable us to obtain an appropriate delay for each time series, facilitating the attainment of significant MP for a specified homological dimension.  

Lastly, we let skip equal 5.  We choose this value for the skip parameter mainly to reach a satisfactory computation time.  The impact of the skip parameter in TDE on MP and computation time is further shown in Supplementary Sec.~\ref{subsec:skip} below.  

Once we set the parameters, the time series are transformed into point clouds.  If the number $|P|$ of points in a point cloud (to be distinguished from the number $|S|$ above of points sampled from the time series prior to TDE) is less than 40, we consider this sample to be invalid, with too few points to represent the original structure of the time series.  Note that this necessary threshold applies to both training and test data for machine learning (cf.~Fig.~\ref{fig:topcap}e).

\subsection{Computing and vectorising persistent homology}

Using Ripser \cite{Sctralie2018ripser,SBauer2021Ripser} for PH (see Supplementary Sec.~\ref{sec:ph} and Introduction in the main text), we compute the PDs of the point clouds in a fast and efficient way.  We then extract MP from each 1-dimensional PD, using persistence birth time and lifetime as two features of a time series.  

The process of vectorising a PD presents a challenge due to the indeterminate (and potentially large) number of intervals in the barcode, coupled with the ambiguous information they contain.  This ambiguity arises from our lack of knowledge about the types of information that can be derived from different parts of the PD.  

Here, we only extract the MP and corresponding birth time.  This decision is informed by our prior selection of an appropriate set of parameters for TDE, which ensures that the MP reaches its theoretical optimum (see Discussion for discussion and experiments beyond optimising MP theoretically).

\section{Discussion on parameter selection}

Here, supplementing Discussion in the main text and Supplementary Sec.~\ref{subsec:m_tde}, we provide further details for a more balanced, comprehensive, and justified perspective.  In the studies applying topological methods to analyse time series \cite{Sbrown_nonlinear_2009,SSP1,STS4,SML4_2,STS5,STS1,STS6}, the determination of parameters for TDE emerges as a pivotal aspect.  This stems from the significant impact that the selection of parameters has on the resulting topological spaces and their corresponding PDs.  There exist several convenient algorithms for parameter selection.  For example, the False Nearest Neighbours algorithm, a widely utilised tool, provides a method for deciding the minimal embedding dimension \cite{Skantz_schreiber_2003}.  However, in the context of PH, usually the objective is not to achieve a minimal dimension.  Contrarily, a dimension of substantial magnitude may be desirable due to certain advantages it offers.

\subsection{Embedding dimension and maximal persistence}
\label{subsec:dim}

In the TDE--PH approach, the determination of dimension in a TDE can be complex.  However, it plays a pivotal role in the extraction of topological descriptors such as MP.  It is observed that a larger dimension can significantly enhance the theoretically optimal MP of a time series.  In TopCap, the dimension of TDE is set to be 100, a relatively large dimension for the experiment.  On the other hand, several factors also constrain this choice.  These include the length of the sampled time series, since the dimension cannot exceed the length, otherwise it would render the resulting point cloud literally pointless.  The constraints also include the periodicity of the time series, as the time-delay window size should be compatible with the approximate period of the time series, at least theoretically as in Supplementary Equation \eqref{tau}.  

According to Perea and Harer \cite[Proposition 5.1]{Sperea_sliding_2015}, there is no information loss for trigonometric polynomials if and only if the dimension of TDE exceeds twice the maximal frequency.  Here, no information loss implies that the original time series can be fully reconstructed from the embedded point cloud.  In general, for a periodic function, a higher dimension of TDE can yield a more precise approximation by trigonometric polynomials.  Although there are no absolutely periodic functions in real data, each time series exhibits its own pattern of vibration, as discussed in the last part of Results in the main text, and a higher dimension of embedding may be employed to capture a more accurate vibration pattern in the time series.  

Moreover, an increased embedding dimension may result in reduced computation time for PD.  For instance, computation times for the voiced consonant [\textipa{N}] in Fig.~\ref{fig:disc}a are 0.2671, 0.2473, and 0.2375 seconds, corresponding to embedding dimensions 10, 100, and 1000.  This is attributed to the reduction on the number of points in the embedded point cloud due to a higher dimension (cf.~CTDE from Discussion in the main text and Supplementary Sec.~\ref{subsec:ctde}).  While this reduction in computation time may not be considered substantial compared to the impact of changing skip (see Supplementary Sec.~\ref{subsec:skip} below), it may become significant when handling large datasets.  

More importantly, an increased embedding dimension can yield benefits such as enhanced MP (cf.~Fig.~\ref{fig:disc}c), which serves as a major motivation for our choice of higher dimensions, as well as a smoother shape of resulting point clouds obtained through TDE, which makes the embedding visibly reasonable.  

Typically, for most algorithms, a lower dimension is preferred due to factors such as those associated with the curse of dimensionality and computational cost.  By contrast, in TopCap, we opt instead for a higher dimension.  

However, the embedding dimension cannot be arbitrarily large.  As illustrated in Supplementary Fig.~\ref{fig:tde}c, when the embedding dimension escalates to 1270, it becomes unfeasible to capture a significant MP in the phonetic time series.  This results from a break of the point cloud.  When the embedding dimension further reaches 1280, an empty 1-dimensional barcode is obtained due to the lack of points necessary to form even a single cycle.  In this way, the dimension of TDE is related to the length of the time series.

\subsection{Skip, maximal persistence, and computation time}
\label{subsec:skip}

\begin{figure*}
    \centering
    \includegraphics[scale=.6]{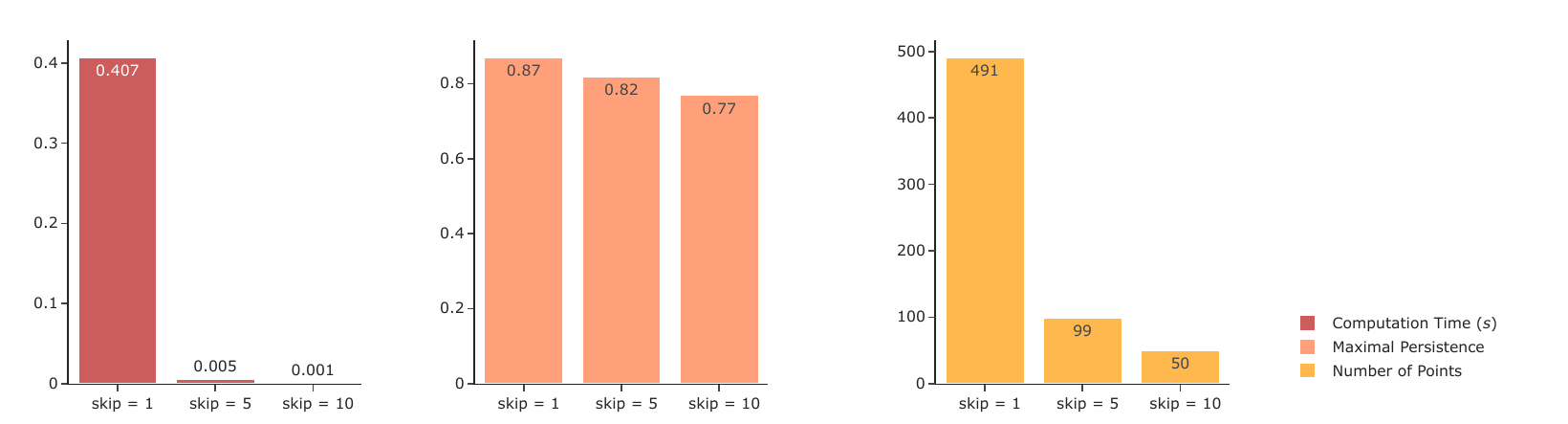}
    \caption{
        \textbf{Computation duration, MP, and the size of point clouds as the skip parameter increases.}  
        An increase in skip can lead to a significant reduction in computation time, owing to the reduced size of the point cloud.  However, MP remains relatively stable.  \\
    }
    \label{fig:skip}
\end{figure*}

\begin{table*}
    \centering
    \begin{tabular}{ccc|ccc|ccc}
        \toprule
        & dimension = 10 & & & dimension = 50 & & & dimension = 100 & \\
        & theoretical delay = 40 & & & theoretical delay = 8 & & & theoretical delay = 4 & \\
        \midrule
        delay & skip & MP & delay & skip & MP & delay & skip & MP \\
        1  & 1 & 0.0610 & 1  & 1 & 0.2834 & 1 & 1 & 0.4270 \\
        10 & 1 & 0.1299 & 3  & 1 & 0.3021 & 2 & 1 & 0.4337 \\
        20 & 1 & 0.1312 & 4  & 1 & 0.3054 & 2 & 5 & 0.4146 \\
        30 & 1 & 0.1281 & 5  & 1 & 0.3058 & 3 & 1 & 0.4357 \\
        39 & 1 & 0.1229 & 6  & 1 & 0.3042 & 3 & 5 & 0.4120 \\
        39 & 5 & 0.1134 & 7  & 1 & 0.3052 & 4 & 1 & 0.4381 \\
        40 & 1 & 0.1290 & 7  & 5 & 0.2886 & 4 & 5 & 0.4139 \\
        40 & 5 & 0.1195 & 8  & 1 & 0.3093 & 5 & 1 & 0.4375 \\
        41 & 1 & 0.1200 & 8  & 5 & 0.2928 & 5 & 5 & 0.4105 \\
        41 & 5 & 0.1153 & 9  & 1 & 0.3091 & 6 & 1 & 0.4347 \\
        45 & 1 & 0.0940 & 9  & 5 & 0.2913 & 6 & 5 & 0.4114 \\
        50 & 1 & 0.1226 & 10 & 1 & 0.3069 & 7 & 1 & 0.4380 \\
        60 & 1 & 0.1315 & 15 & 1 & 0.3070 & 8 & 1 & 0.4378 \\
        94 & 1 & empty  & 18 & 1 & empty  & 9 & 1 & empty  \\
        \bottomrule
    \end{tabular}
    \caption{
        \textbf{1-Dimensional MP for choices of dimension, delay, and skip in TDE.}  
        The theoretical delay is computed as in Supplementary Sec.~\ref{sec:method}.  Empty in MP means that the delay is too large to obtain point-cloud data with proper topological features.  
    }
    \label{tab:parameters}
\end{table*}

Computation time assumes a critical role when a computer processes a substantial volume of data.  In this context, the parameter skip in TDE significantly influences the number of points within the point clouds, and consequently the number of simplices during persistence filtration and the computation time for PD.  

Here, we demonstrate that an appropriate increment in the skip parameter can markedly reduce computation time.  Meanwhile, it is noteworthy that MP exhibits stability with respect to an increase in skip.  As a result, it is feasible to augment skip in TDE to expedite the computation of PD.  On the other hand, for details on the complexity of computing PH, the interested reader may refer to \cite[Sec.~4]{Sedelsbrunner_topological_2002} and \cite[Sec.~4.3]{Szomorodian_computing_2005}.  

Specifically, using an example with a record of the voiced consonant [m], we elucidate the relationship between skip, computation duration, and size of the resulting point clouds obtained via TDE in Supplementary Fig.~\ref{fig:skip}.  Computation duration is measured each time after restarting the Jupyter notebook, on Dell Precision 3581, with CPU Intel$^{\circledR}$ Core$^\text{TM}$ i7-13800H of basic frequency 2.50 GHz and 14 cores.  Computation time means the time for executing the code \texttt{ripser(Points,maxdim=1)}.  As depicted in Supplementary Fig.~\ref{fig:skip}, a substantial reduction in computation time is observed with an increase in the skip parameter.  In contrast, our computation's output MP appears stable.

\subsection{Multiple dependency of maximal persistence}
\label{subsec:mp}

As discussed above, there are three key parameters in TDE, namely, embedding dimension $d$, delay $\tau$, and skip.  In this subsection, we present a table that delineates the topological descriptor MP in relation to these from TopCap.  

The experiment is performed on a record of the voiced consonant [\textipa{N}], which comprises 887 sampled points as the length of this time series.  Theoretically, given a periodic function, one obtains the optimal MP of the function in a fixed dimension under the condition that the TDE window size (i.e., the product of dimension and delay) equals a period (see Supplementary Sec.~\ref{subsec:cts}).  However, the real-world phonetic time series are far from being periodic.  Despite our approach to calculating the period of a time series by ACL functions, we cannot guarantee that the theoretically derived delay will indeed yield the optimal MP of a time series in general.  Nevertheless, this desired delay usually gives relatively good MP.  

For instance, as illustrated in Supplementary Table \ref{tab:parameters}, when the dimension is 10, the theoretical delay is 40.  This corresponds to an MP of 0.1290, which is marginally lower than the MP of 0.1315 achieved at a delay of 60 (cf.~Fig.~\ref{fig:disc}c).  However, as the dimension increases, the point clouds from TDE become more regular.  It becomes increasingly probable that at the desired delay, one can indeed obtain the optimal MP of the time series.  For example, when the dimension is either 50 or 100, the MP of the time series is achieved at the desired delay.  This provides additional justification for preference with higher dimensions.  Indeed, the table reveals that an augmentation in dimension may lead to a more substantial enhancement in the MP of a time series than simply tuning delay.

\section*{References}

\printbibliography[keyword=web,title={\hskip5cm}]

\end{document}